%% file: arxiv.tex
\documentclass[runningheads]{llncs}
\usepackage{graphicx}

\usepackage{tikz}
\usepackage{comment}
\usepackage{amsmath,amssymb} 
\usepackage{color}

\usepackage[accsupp]{axessibility}  

\usepackage{cite}
\usepackage{graphicx}
\usepackage{amsmath}
\usepackage{amssymb}
\usepackage{mathtools}
\usepackage{booktabs}
\usepackage[hashEnumerators,hybrid,underscores=false]{markdown}
\usepackage{xcolor}
\usepackage[utf8x]{inputenc}
\usepackage{pgf}
\usepackage{colortbl}
\usepackage{tikz}
\usepackage{placeins}
\usepackage{tabularx}
\usepackage{todonotes}

\usepackage[pagebackref,breaklinks,colorlinks]{hyperref}
\usepackage{orcidlink}
\usepackage[all]{hypcap}

\newcommand*{\cm}{\checkmark}
\newcommand{\spm}[1]{\tiny{$\,\pm$#1}}

\DeclareMathOperator*{\argmax}{arg\,max}

\newcolumntype{Y}{>{\centering\arraybackslash}X}

\begin{document}
\pagestyle{headings}
\mainmatter
\def\ECCVSubNumber{1999}  

\title{HRDA: Context-Aware High-Resolution Domain-Adaptive Semantic Segmentation} 

\titlerunning{HRDA: Context-Aware High-Res. Domain-Adaptive Semantic Segmentation}
\author{Lukas Hoyer\inst{1}\orcidlink{0000-0002-7391-0676} \and
Dengxin Dai\inst{2}\orcidlink{0000-0001-5440-9678} \and
Luc Van Gool\inst{1,3}\orcidlink{0000-0002-3445-5711}\index{Van Gool, Luc}}
\authorrunning{L. Hoyer et al.}
\institute{ %
ETH Zurich, Switzerland
\email{\{lhoyer,vangool\}@vision.ee.ethz.ch}
\and MPI for Informatics, Germany \email{ddai@mpi-inf.mpg.de} \and KU Leuven, Belgium}
\maketitle

\input{tex_abstract}
\input{tex_introduction}
\input{tex_related_work}
\input{tex_preliminary}

\input{tex_methods}

\input{tex_experiments}
\input{tex_conclusions}

\bibliographystyle{splncs04}
\bibliography{egbib}

\clearpage
\noindent\textbf{\Large Supplementary Material}

\makeatletter
\renewcommand{\thesection}{\Alph{section}}
\renewcommand{\theHsection}{\Alph{section}} 
\renewcommand{\thefigure}{S\arabic{figure}}
\renewcommand{\theHfigure}{S\arabic{figure}}
\renewcommand{\thetable}{S\arabic{table}}
\renewcommand{\theHtable}{S\arabic{table}}
\setcounter{section}{0}
\setcounter{figure}{0}
\setcounter{table}{0}
\makeatother

\input{tex_supplement}

\end{document}

%% file: tex_abstract.tex
\begin{abstract}
Unsupervised domain adaptation (UDA) aims to adapt a model trained on the source domain (e.g. synthetic data) to the target domain (e.g. real-world data) without requiring further annotations on the target domain. This work focuses on UDA for semantic segmentation as real-world pixel-wise annotations are particularly expensive to acquire. As UDA methods for semantic segmentation are usually GPU memory intensive, most previous methods operate only on downscaled images. We question this design as low-resolution predictions often fail to preserve fine details. The alternative of training with random crops of high-resolution images alleviates this problem but falls short in capturing long-range, domain-robust context information. Therefore, we propose HRDA, a multi-resolution training approach for UDA, that combines the strengths of small high-resolution crops to preserve fine segmentation details and large low-resolution crops to capture long-range context dependencies with a learned scale attention, while maintaining a manageable GPU memory footprint. HRDA enables adapting small objects and preserving fine segmentation details. It significantly improves the state-of-the-art performance by 5.5 mIoU for GTA$\rightarrow$Cityscapes and 4.9 mIoU for Synthia$\rightarrow$Cityscapes, resulting in unprecedented 73.8 and 65.8 mIoU, respectively. The implementation is available at \texttt{\href{https://github.com/lhoyer/HRDA}{github.com/lhoyer/HRDA}}.

\keywords{Unsupervised Domain Adaptation; Semantic Segmentation; Multi-Resolution; High-Resolution; Attention}
\end{abstract}

%% file: tex_introduction.tex
\section{Introduction}

Even though neural networks currently are the unchallenged approach to solve many computer vision problems, their training often requires a large amount of annotated data. For certain tasks, such as semantic segmentation, providing the annotations is particularly labor-intensive as pixel-wise labels of the entire image are necessary, which can take more than one hour per image~\cite{cordts2016cityscapes, sarkadis2021acdc}.
Therefore, several methods aim to reduce the annotation burden such as weakly-supervised learning~\cite{dai2015boxsup, zou2021pseudoseg, unal2022scribble}, semi-supervised learning~\cite{souly2017semi, french2019consistency, hoyer2021three, lai2021semi}, and unsupervised domain adaption (UDA)~\cite{hoffman2016fcns, tsai2018learning, zou2018unsupervised,  hoyer2021daformer, wang2022continual}. 
In this work, we focus on UDA. To avoid the annotation effort for the target dataset, the network is trained on a source dataset with existing or cheaper annotations such as automatically labeled synthetic data~\cite{ros2016synthia, richter2016playing}. However, neural networks are usually sensitive to domain shifts. This problem is approached in UDA by adapting the network, which is trained with source data, to unlabeled target images.

\begin{figure}[t]
\centering
\includegraphics[width=0.73\linewidth]{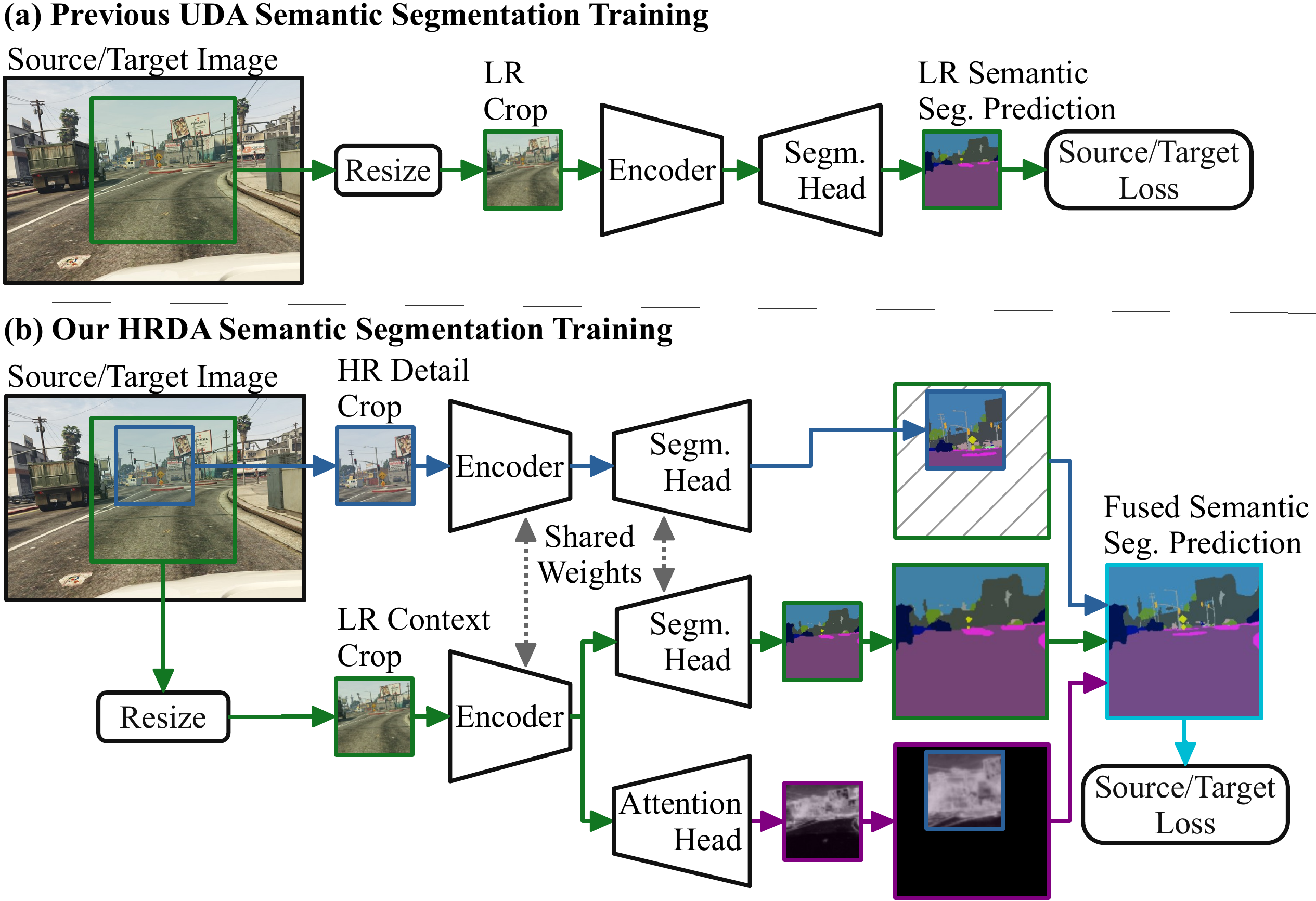}
\resizebox{0.26\linewidth}{!}{\input{fig/uda_over_time_hrda.pgf}}
\caption{(a) Most previous UDA methods were only trained with downscaled inputs to account for their high GPU memory footprint. (b) Our HRDA incorporates fine segmentation details from a random high-resolution (HR) detail crop and context information from a random low-resolution (LR) context crop. Their predictions are fused using a learned scale attention (best viewed zoomed-in). In that way, HRDA can utilize HR details and long-range context information while keeping a manageable memory footprint. (c) Compared to previous works, HRDA provides a major performance gain.}
\label{fig:overview}
\end{figure}

UDA methods are usually more GPU memory intensive than regular supervised learning as UDA training often requires images from multiple domains, additional networks (e.g. teacher model or domain discriminator), and additional losses, which consume significant additional GPU memory.
Therefore, most UDA semantic segmentation methods so far (e.g.~\cite{tsai2018learning, zhou2017unsupervised, yang2020fda, tranheden2021dacs, araslanov2021self, wang2021domain, liu2021bapa, hoyer2021daformer}) follow the convention of downscaling images due to GPU memory constraints (see Fig.~\ref{fig:overview} a).
Taking Cityscapes as an example, current UDA methods use half the full resolution (i.e. $1024{\times}512$ pixels), while most supervised methods use the full resolution (i.e. $2048{\times}1024$ pixels).
This is one of the key differences in the training setting of UDA and supervised semantic segmentation, possibly contributing to the gap between the state-of-the-art performance of UDA and supervised learning.

We question this design choice as predictions from low-resolution (LR) inputs often fail to recognize small objects such as distant traffic lights and to preserve fine segmentation details such as limbs of distant pedestrians.
However, naively learning UDA with full high-resolution (HR) images is often difficult as the resolution quadratically affects the GPU memory consumption. A common remedy is training with random crops of the image. While training with HR crops allows to adapt small objects and preserve segmentation details, it limits the learned long-range context dependencies to the crop size. This is a crucial disadvantage for UDA as context information and scene layout are often domain-robust (e.g. rider on bicycle or sidewalk at the side of road)~\cite{huang2020contextual, yang2021context, zhou2021context}. Further, while HR inputs are necessary to adapt small objects, they can be disadvantageous compared to LR inputs when adapting large stuff-regions such as close sidewalks (see Sec.~\ref{sec:exp_resolution_crop_size}).
At HR, these regions often contain too detailed, domain-specific features (e.g. detailed sidewalk texture), which are detrimental to UDA. LR inputs `hide away' these features, while still providing sufficient details to recognize large regions across domains.

To effectively combine the strength of these two approaches while maintaining a manageable GPU memory footprint, we propose \emph{HRDA}, a novel multi-resolution framework for UDA semantic segmentation  (see Fig.~\ref{fig:overview} b).
First, HRDA uses a large LR \emph{context crop} to adapt large objects without confusion from domain-specific HR textures and to learn long-range context dependencies as we assume that HR details are not crucial for long-range dependencies.
Second, HRDA uses a small HR \emph{detail crop} from the region within the context crop to adapt small objects and to preserve segmentation details as we assume that long-range context information play only a subordinate role in learning segmentation details. In that way, the GPU memory consumption is significantly reduced while still preserving the main advantages of a large crop size and a high resolution. 
Given that the importance of the LR context crop vs. the HR detail crop depends on the content of the image, HRDA fuses both using an input-dependent scale attention. During UDA, the attention learns to decide how trustworthy the LR and the HR predictions are in every image region.
Previous multi-resolution frameworks for supervised learning~\cite{chen2016attention, yang2018attention, tao2020hierarchical} cannot naively be applied to state-of-the-art UDA due to GPU memory constraints as they operate on full LR and HR images.
To adapt HRDA to the target domain, it can be trained with pseudo-labels fused from multiple resolutions. To further increase the robustness of the detail pseudo-labels with respect to different contexts, they are generated using an overlapping sliding window mechanism.

This work makes four contributions. To the best of our knowledge, it is the first work on UDA semantic segmentation (1) systematically studying the influence of resolution and crop size, (2) exploiting HR inputs for adapting small objects and fine segmentation details, (3) applying multi-resolution fusion with a learned scale attention for object-scale-dependent adaptation, and 
(4) fusing a nested large LR crop to capture long-range context and small HR crop to capture details for memory-efficient UDA training.
HRDA provides significant performance gains when applied to various UDA strategies~\cite{tsai2018learning, vu2019advent, tranheden2021dacs, hoyer2021daformer}. When used with the SOTA method DAFormer~\cite{hoyer2021daformer}, HRDA gains +5.5 mIoU for GTA$\rightarrow$Cityscapes and +4.9 mIoU for Synthia$\rightarrow$City\-scapes, resulting in unprecedented 73.8 and 65.8 mIoU, respectively (see Fig.~\ref{fig:overview} c).

%% file: tex_related_work.tex
\section{Related Work}

\noindent\textbf{Semantic Segmentation:}
Most semantic segmentation approaches are based on (convolutional) neural networks, which can be effectively trained in an end-to-end manner to perform pixel-wise classification as first shown by Long et al.~\cite{long2015fully}. This concept was further improved in different aspects including increasing the receptive field while preserving spatial details~\cite{chen2014semantic,ronneberger2015u,yu2015multi,zhao2017pyramid,chen2017deeplab,chen2018encoder,wang2020deep}, integrating context information~\cite{zhang2018context,zhou2019context,hoyer2019grid,yuan2020object,yuan2021ocnet}, attention mechanisms~\cite{wang2018non,zhao2018psanet,fu2019dual,huang2019ccnet}, refining boundaries~\cite{ding2019boundary,li2020improving,yuan2020segfix}, 
and Transformer-based architectures~\cite{zheng2021rethinking,xie2021segformer,wang2021pyramid,strudel2021segmenter,liu2021swin}.

Several architectures~\cite{zhao2017pyramid,chen2017deeplab,chen2017rethinking,ding2019acnet,lin2019zigzagnet} utilize intermediate features with different scales, which are generated from a single scale input, to aggregate context information. 
Furthermore, multi-scale input inference, where predictions from scaled versions of an image are combined via average or max pooling, is often used to obtain better results~\cite{chen2017rethinking,chen2018encoder,cheng2020panoptic,yuan2020object}. However, the naive pooling is independent of the image content, which can lead to suboptimal results. Therefore, Chen et al.~\cite{chen2016attention} and Yang et al.~\cite{yang2018attention} segment multi-scale inputs and learn an attention-weighted sum of the predictions. Tao et al.~\cite{tao2020hierarchical} further propose a hierarchical attention that is agnostic to the number of scales during inference. 

\noindent\textbf{Unsupervised Domain Adaptation (UDA):}
To adapt a semantic segmentation network to the target domain, several strategies have been proposed, while most can be grouped into adversarial training and self-training. In adversarial training~\cite{goodfellow2014generative,ganin2016domain}, a domain discriminator is trained in order to provide supervision to align the domain distributions based on style-transferred inputs~\cite{hoffman2018cycada, li2019bidirectional, pizzati2020domain, gong2021dlow} or network features/outputs~\cite{hoffman2016fcns, tsai2018learning, tsai2019domain, vu2019advent, wang2020classes}. In self-training, the network is adapted to the target domain using high-confidence pseudo-labels. In order to regularize the training and to avoid pseudo-label drift, approaches such as confidence thresholding~\cite{zou2018unsupervised,mei2020instance}, pseudo-label prototypes~\cite{zhang2019category,zhang2021prototypical,liu2021bapa}, and consistency regularization~\cite{sajjadi2016regularization, tarvainen2017mean, sohn2020fixmatch} based on data augmentation~\cite{choi2019self, melaskyriazi2021pixmatch, araslanov2021self, prabhu2022augco}, different context~\cite{lai2021semi,zhou2021context}, domain-mixup~\cite{tranheden2021dacs, zhou2021context,liu2021bapa,gao2021dsp,hoyer2021improving}, or multiple models~\cite{zhou2020uncertainty,zheng2021rectifying,zhang2021multiple} have been used. Several works also combine self-training and adversarial training~\cite{li2019bidirectional, wang2020classes, kim2020learning, zheng2020unsupervised}, minimize the entropy~\cite{vu2019advent,vu2019dada,chen2019domain}, refine boundaries~\cite{liu2021bapa}, use a curriculum~\cite{dai2018dark,zhang2019curriculum,dai2020curriculum}, or learn auxiliary tasks~\cite{vu2019dada,chen2019learning,wang2021domain,hoyer2021improving}.
The use of semantic segmentation networks with multi-scale \emph{features} is quite common in UDA as many methods evaluate their approach with DeepLabV2~\cite{chen2017deeplab}. However, these features are generated from a single-scale input.
While some works apply multi-scale average pooling for inference~\cite{araslanov2021self, wang2020classes} or enforce scale consistency~\cite{subhani2020learning,iqbal2020mlsl,guan2021scale} of low-resolution inputs, they fall short in learning an input-adaptive multi-scale fusion.
To the best of our knowledge, HRDA is the first work to learn a multi-resolution \emph{input} fusion for UDA semantic segmentation.
For that purpose, we newly extend scale attention~\cite{chen2016attention,tao2020hierarchical} to UDA and reveal its significance for UDA by improving the adaptation process across different object scales.
Further, we propose fusing nested crops with different scales and sizes, which successfully overcomes the pressing issue of limited GPU memory for multi-resolution UDA.

%% file: tex_preliminary.tex
\section{Preliminary}
\label{sec:preliminary}

In UDA, a neural network $f_\theta$ is trained using source domain images $\mathcal{X}^S = \{x_{\mathit{HR}}^{S,m}\}_{m=1}^{N_S}$ with $x_\mathit{HR}^{S,m} \in \mathbb{R}^{H_S \times W_S \times 3}$ and target domain images $\mathcal{X}^T= \{x_{\mathit{HR}}^{T,m}\}_{m=1}^{N_T}$ with $x_\mathit{HR}^{T,m} \in \mathbb{R}^{H_T \times W_T \times 3}$ to achieve a good performance on the target domain. However, labels are only available for the source domain $\mathcal{Y}^S = \{y_\mathit{HR}^{S,m}\}_{m=1}^{N_S}$ with $y_\mathit{HR}^{S,m} \in \{0,1\}^{H_S \times W_S \times C}$. As the following definitions refer to the same source/target sample, we will drop index $m$ to avoid convolution.
Most previous UDA methods bilinearly downsample $\zeta(\cdot, \cdot)$ the images and labels $x_\mathit{HR}^S$, $x_\mathit{HR}^T$, and $y_\mathit{HR}^S$ by a dataset-specific factor $s_S,s_T \geq 1$ to satisfy GPU memory constraints, e.g. $x_\mathit{LR}^T = \zeta(x_\mathit{HR}^T, 1 / s_T)  \in \mathbb{R}^{\frac{H_T}{s_T} \times \frac{W_T}{s_T} \times 3}$.
Some methods such as \cite{tranheden2021dacs,zhang2021prototypical,hoyer2021daformer} additionally crop the LR image but for simplicity, we consider full LR images in this section.

As only source labels are available, the supervised categorical cross-entropy loss can only be calculated for the source predictions $\hat{y}_\mathit{LR}^S = f_\theta(x_\mathit{LR}^S)$
\begin{align}
    \mathcal{L}^S &= \mathcal{L}_\mathit{ce}(\hat{y}_\mathit{LR}^S, y_\mathit{LR}^S, 1)\,,\\
    \mathcal{L}_\mathit{ce}(\hat{y}, y, q) &= -\sum_{i=1}^{H(y)} \sum_{j=1}^{W(y)} \sum_{c=1}^C q_{ij} y_{ijc} \log \zeta(\hat{y}, \frac{H(y)}{H(\hat{y})})_{ijc}\,.
\end{align}
As the predictions are usually smaller than the input due to the output stride of the segmentation network, they are upsampled to the label size $H(y) \times W(y)$.

However, a model trained only with $\mathcal{L}^S$ usually does not generalize well to the target domain. In order to adapt the model to the target domain, UDA methods incorporate an additional loss for the target domain $\mathcal{L}^T$, which is added to the overall loss $\mathcal{L} = \mathcal{L}^S + \lambda \mathcal{L}^T$.
The target loss can be defined according to the used training strategies such as adversarial training~\cite{tsai2018learning,tsai2019domain,wang2020classes} or self-training~\cite{zou2018unsupervised,zhang2019category,mei2020instance,tranheden2021dacs,zhang2021prototypical,hoyer2021daformer}. In this work, we mainly evaluate HRDA with the self-training method DAFormer~\cite{hoyer2021daformer}, as it is the current state-of-the-art method for UDA semantic segmentation.
In self-training, the model is iteratively adapted to the target domain by training it with pseudo-labels for target images, predicted by a teacher network $g_\phi$:
\begin{equation}
    p_\mathit{LR,ijc}^T = [c = \argmax_{c'} g_\phi(x_\mathit{LR}^T)_{ijc'}]\,,
\end{equation}
where $[\cdot]$ denotes the Iverson bracket. The pseudo-labels are used to additionally train the network $f_\theta$ on the target domain
\begin{equation}
    \mathcal{L}^T = \mathcal{L}_\mathit{ce}(\hat{y}_\mathit{LR}^T, p_\mathit{LR}^T, q_\mathit{LR}^T)\,.
\end{equation}
As the pseudo-labels are not necessarily correct, their quality is weighted by a confidence estimate $q_\mathit{LR}^T$~\cite{zou2018unsupervised,mei2020instance,tranheden2021dacs,hoyer2021daformer}. 
After each training step $t$, the teacher model $g_\phi$ is updated with the exponentially moving average of the weights of $f_\theta$, implementing a temporal ensemble to stabilize pseudo-labels, which is a common strategy in semi-supervised learning~\cite{tarvainen2017mean,sohn2020fixmatch,french2019consistency} and UDA~\cite{araslanov2021self,tranheden2021dacs,liu2021bapa}
\begin{equation}
    \phi_{t+1} \leftarrow \alpha \phi_t + (1 - \alpha) \theta_t\,.
\end{equation}

Further, DAFormer~\cite{hoyer2021daformer} uses consistency training~\cite{sajjadi2016regularization, tarvainen2017mean, sohn2020fixmatch}, i.e. the network $f_\theta$ is trained on augmented target data following DACS~\cite{tranheden2021dacs}, while the teacher model $g_\phi$ generates the pseudo-labels using non-augmented target images.
Besides self-training, DAFormer~\cite{hoyer2021daformer} uses a domain-robust Transformer network, rare class sampling, and feature regularization based on ImageNet features.

%% file: tex_methods.tex
\section{Methods}

In this work, we propose a multi-resolution framework for UDA as small objects and segmentation details are easier to adapt with high-resolution (HR) inputs, while large stuff regions are easier to adapt with low-resolution (LR) inputs.
As UDA methods require more GPU memory than regular supervised training, we design a training strategy based on a large LR context crop to learn long-range context dependencies and a small HR detail crop to preserve segmentation details while maintaining a manageable GPU memory footprint (Sec.~\ref{sec:context_detail_crop}). The strengths of both LR context and HR detail crop are combined by fusing their predictions with a learned scale attention (Sec.~\ref{sec:multi_resolution_uda}).
For a robust pseudo-label generation, we further utilize overlapping slide inference to fuse predictions with different contexts (Sec.~\ref{sec:overlapping_slide_inference}).
The proposed method is designed to be applicable to common network architectures and can be combined with existing UDA methods.

\subsection{Context and Detail Crop}
\label{sec:context_detail_crop}

\begin{figure}
\centering
\includegraphics[width=\linewidth]{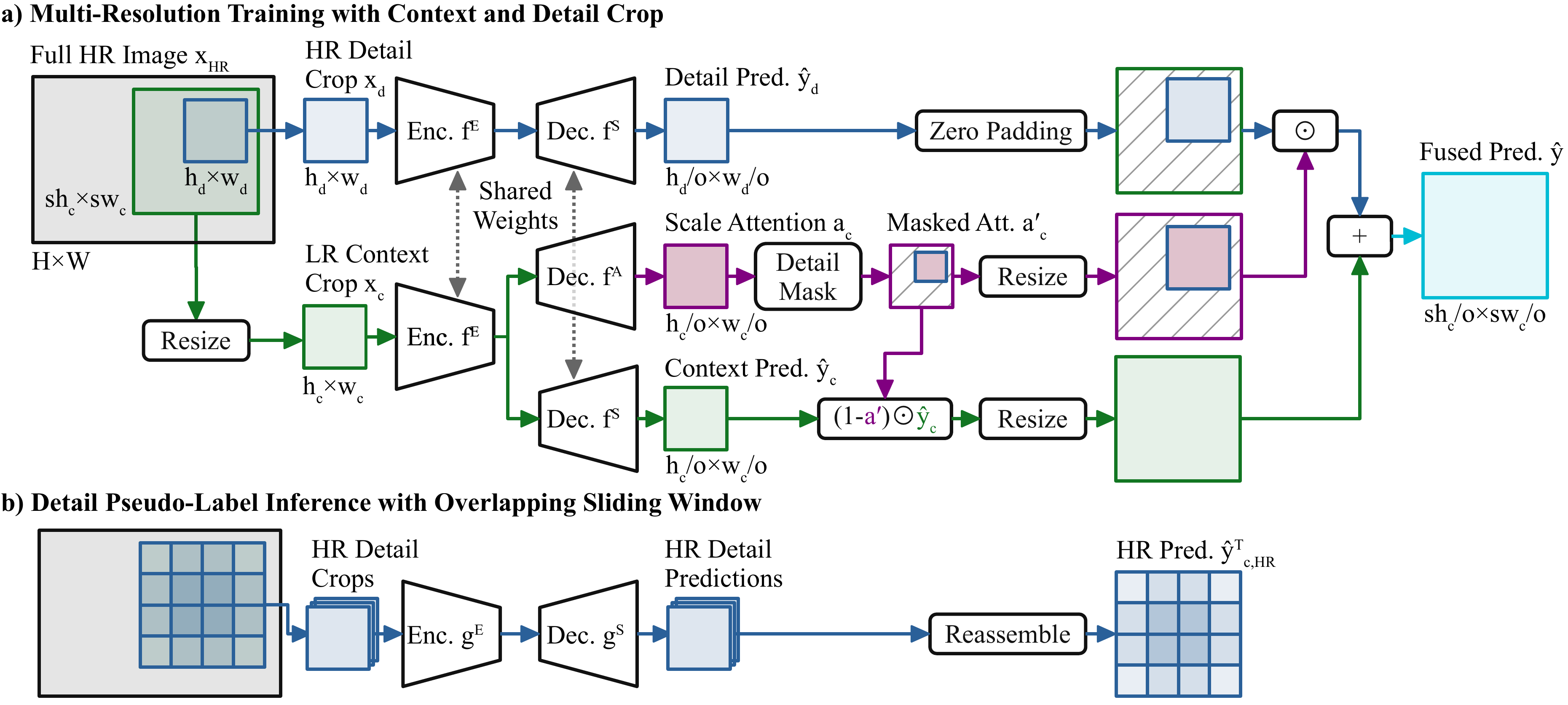}
\caption{(a) Multi-resolution training with low-resolution (LR) context and high-resolution (HR) detail crop. The prediction of the detail crop is fused into the context prediction within the region where it was cropped from by a learned scale attention. (b) For pseudo-label generation, multiple detail crops are generated using overlapping slide inference to cover the entire context crop. The pseudo-label is fused from HR pred. $\hat{y}_\mathit{c,HR}^T$ and LR pred. $\hat{y}_\mathit{c}^T$ with the full attention $a_c^T$ similar to (a) (see Sec.~\ref{sec:overlapping_slide_inference}). 
}
\label{fig:architecture}
\end{figure}

Due to GPU memory constraints, it is not feasible to train state-of-the-art UDA methods with full-sized high-/multi-resolution inputs as images from multiple domains, additional networks, and additional losses are required for UDA training. Therefore, most previous works only use LR inputs. However, HR inputs are important to recognize small objects and produce fine segmentation borders. In order to still utilize HR inputs, random cropping is a possible solution. However, random cropping restricts learning context-aware semantic segmentation, especially for long-range dependencies and scene layout, which might be critical for UDA as context relations are often domain-invariant (e.g. car on road, rider on bicycle)~\cite{huang2020contextual, yang2021context, zhou2021context}. In order to both train with long-range context as well as high resolution, we propose to combine different crop sizes for different resolutions, i.e. a large LR context crop and a small HR detail crop (see Fig.~\ref{fig:architecture} a). The purpose of the context crop is to provide a large crop to learn long-range context relations. The purpose of detail crop is to focus on HR to recognize small objects and produce fine segmentation details, which does not necessarily require far-away context information.
In order to segment the entire image during model validation, overlapping sliding window inference is used (see Sec.~\ref{sec:overlapping_slide_inference}).

The context crop $x_c \in \mathbb{R}^{h_c \times w_c \times 3}$ is obtained by cropping from the original HR image $x_\mathit{HR} \in \mathbb{R}^{H \times W \times 3}$ and bilinear downsampling by the factor $s \geq 1$
\begin{equation}
    x_{\mathit{c,HR}} = x_\mathit{HR}[b_{c,1} : b_{c,2}, b_{c,3} : b_{c,4}]\,, \quad x_c = \zeta(x_{c,\mathit{HR}}, 1/s)
    \label{eq:x_c}
\end{equation}
The crop bounding box $b_c$ is randomly sampled from a discrete uniform distribution within the image size while ensuring that the coordinates can be divided by $k=s \cdot o$ with $o \geq 1$ denoting the output stride of the segmentation network to ensure exact alignment in the later fusion process:
\begin{alignat}{3}
    b_{c,1} &\sim \mathcal{U}\{0, (H-sh_c) / k\} \cdot k\,,\quad &&b_{c,2} = b_{c,1} + sh_c\,,\\ 
    b_{c,3} &\sim \mathcal{U}\{0, (W-sw_c) / k\} \cdot k\,,\quad &&b_{c,4} = b_{c,3} + sw_c\,.\nonumber
\end{alignat}
The detail crop $x_d \in \mathbb{R}^{h_d \times w_d \times 3}$ is randomly cropped from within the context crop region, which is necessary to enable the fusion of context and detail predictions in the later process:
\begin{alignat}{3}
    x_d &= x_\mathit{c,HR}[b_{d,1} : b_{d,2}, b_{d,3} : b_{d,4}] &&\,,\label{eq:x_d}\\
    b_{d,1} &\sim \mathcal{U}\{0, (sh_c-h_d) / k\} \cdot k\,,\quad &&b_{d,2} &&= b_{d,1} + h_d\,,\\ 
    b_{d,3} &\sim \mathcal{U}\{0, (sw_c-w_d) / k\} \cdot k\,,\quad &&b_{d,4} &&= b_{d,3} + w_d\,.\nonumber
\end{alignat}
In this work, we use context and detail crops of the same dimension, i.e. $h_c=h_d$ and $w_c=w_d$, to balance the required resources for both crops and provide a good trade-off between context-aware and detailed predictions. The context downscale factor is $s=2$ following the LR design of previous UDA methods~\cite{tranheden2021dacs,liu2021bapa,hoyer2021daformer}, which means that the context crop covers 4 times more content at half the resolution compared to the detail crop.

Using a feature encoder $f^E$ and a semantic decoder $f^S$, the context semantic segmentation $\hat{y}_c = f^S(f^E(x_c)) \in \mathbb{R}^{\frac{h_c}{o} \times \frac{w_c}{o} \times C}$ and the detail semantic segmentation $\hat{y}_d = f^S(f^E(x_d)) \in \mathbb{R}^{\frac{h_d}{o} \times \frac{w_d}{o} \times C}$ are predicted. 
The networks $f^E$ and $f^S$ are shared for both HR and LR inputs. 
This not only saves memory usage but also increases the robustness of the network against different resolutions.

\subsection{Multi-Resolution Fusion}
\label{sec:multi_resolution_uda}

While the HR detail crop is well-suited to segment small objects such as poles or distant pedestrians, it lacks the ability to capture long-range dependencies, which is disadvantageous in segmenting large stuff regions such as large regions of sidewalk. The opposite is the case for the LR context crop. Therefore, we fuse the predictions from both crops using a learned scale attention~\cite{chen2016attention} to predict in which image regions to trust predictions from context and detail crop. Additionally, the scale attention provides the advantage that it enables adapting objects at the better-suited scale. For example, small objects are easier to adapt at HR while large objects are easier to adapt at LR as the appearance of an object
should have a resolution high enough to be discriminative but not too high to avoid that the network overfits to domain-specific detailed textures.

The scale attention decoder $f^A$ learns to predict the scale attention $a_c = \sigma(f^A(f^E(x_c))) \in [0,1]^{\frac{h_c}{o} \times \frac{w_c}{o} \times C}$ to weigh the trustworthiness of LR context and HR detail predictions. The sigmoid function $\sigma$ ensures a weight in $[0, 1]$, where 1 means a focus on the HR detail crop. The attention is predicted from the context crop as it has a better grasp on the scene layout (larger context).
As the predictions are smaller than the inputs due to the output stride $o$, the crop coordinates are scaled accordingly in the following steps. Outside of the detail crop $c_d$, the attention is set to 0 as there is no detail prediction available
\begin{equation}
    a'_c \in \mathbb{R}^{\frac{h_c}{o} \times \frac{w_c}{o}}\,,\quad a'_c(i,j) = 
    \begin{cases} 
        a_c(i,j) \quad &\text{if} \, \frac{b_{d,1}}{s \cdot o} \leq i < \frac{b_{d,2}}{s \cdot o} \wedge \frac{b_{d,3}}{s \cdot o} \leq j < \frac{b_{d,4}}{s \cdot o}\\
        0 \quad &\text{otherwise} \\
  \end{cases}\,.
  \label{eq:attention_mask}
\end{equation}
The detail crop is aligned with the (upsampled) context crop by padding it with zeros to a size of $\frac{sh_c}{o} \times \frac{sw_c}{o}$
\begin{equation}
    \hat{y}_d'(i,j) = 
    \begin{cases} 
        \hat{y}_d(i-\frac{b_{d,1}}{o},j-\frac{b_{d,3}}{o}) \quad &\text{if} \, \frac{b_{d,1}}{o} \leq i < \frac{b_{d,2}}{o} \wedge \frac{b_{d,3}}{o} \leq j < \frac{b_{d,4}}{o}\\
        0 \quad &\text{otherwise} \\
  \end{cases}\,.
\end{equation}
The predictions from multiple scales are fused using the attention-weighted sum
\begin{equation}
    \hat{y}_\mathit{c,F} = \zeta((1 - a'_c) \odot \hat{y}_c, s) + \zeta(a'_c, s) \odot \hat{y}_d'\,.
    \label{eq:scale_fusion}
\end{equation}
The encoder, segmentation head, and attention head are trained with the fused multi-scale prediction and the detail crop prediction
\begin{equation}
    \mathcal{L}_\mathit{HRDA}^S = (1 - \lambda_d) \mathcal{L}_\mathit{ce}(\hat{y}_{c,F}^S, y_{c,\mathit{HR}}^S, 1) + \lambda_d \mathcal{L}_\mathit{ce}(\hat{y}_{d}^S, y_{d}^S, 1)\,,
\end{equation}
where the ground truth $y_\mathit{c,HR}^S$/$y_d^S$ is cropped according to Eq.~\ref{eq:x_c}/\ref{eq:x_d}. Additionally supervising the detail crop is helpful to learn more robust features for HR inputs even though the attention might favor the context crop in that region. An additional loss for $\hat{y}_c$ is not necessary as it is already directly supervised in regions without detail crop (see Eq.~\ref{eq:attention_mask}).
The target loss $\mathcal{L}_\mathit{HRDA}^T$ is adapted accordingly
\begin{equation}
    \mathcal{L}_{\mathit{HRDA}}^{T} = 
    (1 - \lambda_d) \mathcal{L}_\mathit{ce}(\hat{y}_{c,F}^{T}, p_{c,F}^{T}, q_{c,F}^{T})
    + \lambda_d \mathcal{L}_\mathit{ce}(\hat{y}_{d}^{T}, p_{d}^{T}, q_{d}^{T})\,.
    \label{eq:L_T,HRDA}
\end{equation}
For pseudo-label prediction, we also utilize multi-resolution fusion (see Sec.~\ref{sec:overlapping_slide_inference}). Therefore, when predicting pseudo-labels the scale attention focuses on the better-suited resolution (e.g. HR for small objects). As the pseudo-labels are further used to train the model also with the worse-suited resolution (e.g. LR for small objects), it improves the robustness for both small and large objects. 

\subsection{Pseudo-Label Generation with Overlapping Sliding Window}
\label{sec:overlapping_slide_inference}

For self-training with Eq.~\ref{eq:L_T,HRDA}, it is necessary to generate a high-quality HRDA pseudo-label $p_\mathit{c,F}^T$ for the context crop $x_\mathit{c,HR}^T$. The underlying HRDA prediction $\hat{y}_\mathit{c,F}^T$ is fused from the LR prediction $\hat{y}_c^T$ and HR prediction $\hat{y}_\mathit{c,HR}^T$ using the full scale attention $a_c^T$ similar to Eq.~\ref{eq:scale_fusion} \begin{equation}
    \hat{y}_\mathit{c,F}^T = \zeta((1 - a_c^T) \odot \hat{y}_c^T, s) + \zeta(a_c^T, s) \odot \hat{y}_\mathit{c,HR}^T\,.
\end{equation}
Therefore, the HR prediction $\hat{y}_\mathit{c,HR}^T$ is necessary for the entire context crop $x_\mathit{c,HR}^T$ instead of just the detail crop $x_d$. Note that for pseudo-label generation $g_\phi$ instead of $f_\theta$ is used for predictions.
Even though large HR network inputs are problematic during training, they are not an issue during pseudo-label inference as no data for the backpropagation has to be stored. 
However, the used DAFormer~\cite{hoyer2021daformer}, as well as other Vision Transformers~\cite{xie2021segformer,zheng2021rethinking}, have a learned (implicit) positional embedding that works best if training and inference input size are the same.
Therefore, we infer the HR prediction $\hat{y}_\mathit{c,HR}^T$ using a sliding window of size $h_d \times w_d$ over the HR context crop $x_\mathit{c,HR}^T$ (see Fig~\ref{fig:architecture} b). The window is shifted with a stride of $h_d/2 \times w_d/2$ to generate overlapping predictions with different contexts, which are averaged to increase robustness.
The crops of the sliding window can be processed in parallel as the images in a batch, which allows for efficient computation on the GPU.

For model validation or deployment, the full-scale HRDA semantic segmentation $\hat{y}_\mathit{F,HR}$ of the entire image $x_\mathit{HR}$ is necessary. As the context crop is usually smaller than the entire image, $\hat{y}_\mathit{F,HR}$ is generated using an overlapping sliding window over the entire image $x_\mathit{HR}$ with a size of $sh_c \times sw_c$ and a stride of $sh_c/2 \times sw_c/2$. Within the sliding window, the HRDA prediction is generated in the same way as $\hat{y}_\mathit{c,F}^T$ for the pseudo-label.

%% file: tex_experiments.tex
\section{Experiments}

\subsection{Implementation Details}

\noindent\textbf{Datasets:}
As target data, the real-world Cityscapes dataset~\cite{cordts2016cityscapes} of European street scenes with 2975 training and 500 validation images of 2048${\times}$1024 pixels is used.
As source data, the synthetic datasets GTA~\cite{richter2016playing} with 24,966 images of 1914$\times$1052 pixels and Synthia~\cite{ros2016synthia} with 9,400 images of 1280$\times$760 pixels are used. 
Previous works~\cite{tsai2018learning,yang2020fda,tranheden2021dacs,hoyer2021daformer} downsample Cityscapes to 1024${\times}$512 and GTA to 1280${\times}$720. Instead, we maintain the full resolution for Cityscapes. To train with the same scale ratio of source and target images as previous works, we resize GTA to 2560${\times}$1440 and Synthia to 2560${\times}$1520 pixels.

\noindent\textbf{Network Architecture:}
Our default network is based on DAFormer~\cite{hoyer2021daformer}. It consists of an MiT-B5 encoder~\cite{xie2021segformer} and a context-aware feature fusion decoder~\cite{hoyer2021daformer}. For the scale attention decoder, we use the lightweight SegFormer MLP decoder~\cite{xie2021segformer} with an embedding dimension of 256.
When evaluating other UDA methods in Tab.~\ref{tab:other_uda}, we use a ResNet101~\cite{he2016deep} backbone with a DeepLabV2~\cite{chen2017deeplab} decoder both as segmentation and scale attention head. 

\noindent\textbf{Training:}
By default, we follow the DAFormer~\cite{hoyer2021daformer} self-training strategy (see Sec.~\ref{sec:preliminary}) and training parameters, i.e. AdamW~\cite{loshchilov2018decoupled} with a learning rate of $6 {\times} 10^{-5}$ for the encoder and $6 {\times} 10^{-4}$ for the decoder, a batch size of 2, linear learning rate warmup, $\lambda_\mathit{st}{=}1$, $\alpha{=}0.999$, and DACS~\cite{tranheden2021dacs} data augmentation. For adversarial training and entropy minimization, we use SGD with a learning rate of $0.0025$ and $\lambda_\mathit{adv}{=}\lambda_\mathit{ent}{=}0.001$. The context and detail crop are generated using $h_c{=}w_c{=}h_d{=}w_d{=}512$ with $s{=}2$ to balance the required resources for both crops in the default case. The detail loss weight is chosen empirically $\lambda_d{=}0.1$. 
The experiments are conducted on a Titan RTX GPU with 24 GB memory.

\subsection{Comparison with State-of-the-Art UDA Methods}

\begin{figure*}[tb]
\centering
\input{preds/prediction_head}
\includegraphics[width=\linewidth]{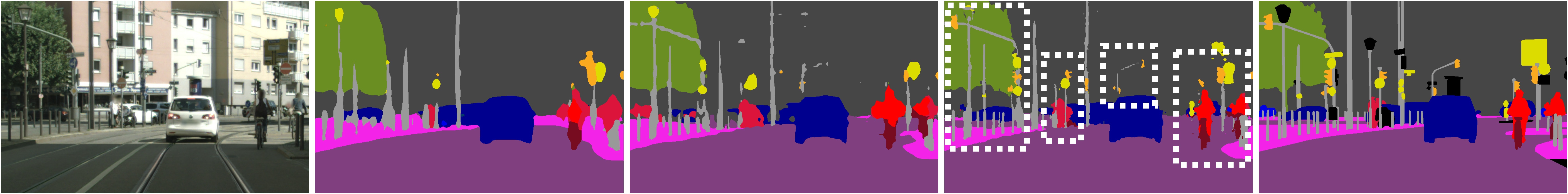}
\includegraphics[width=\linewidth]{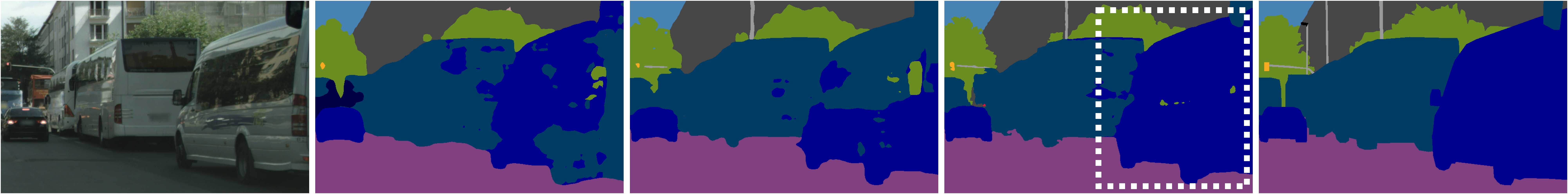}
\caption{Qualitative comparison of HRDA with previous methods on GTA$\rightarrow$Cityscapes. HRDA improves the segmentation of small classes such as pole, traffic sign, traffic light, and rider as well as large and difficult classes such as bus.}
\label{fig:visual_examples}
\end{figure*}

First, we compare the proposed HRDA with previous UDA methods in Tab.~\ref{tab:sota}.
It can be seen that HRDA outperforms the previously best state-of-the-art method by a significant margin of +5.5 mIoU on GTA$\rightarrow$Cityscapes and +4.9 mIoU on Synthia$\rightarrow$ Cityscapes. HRDA improves the IoU of almost all classes across both datasets. The highest performance gains are achieved for classes with fine segmentation details such as pole, traffic light, traffic sign, person, rider, motorbike, and bike. But also large classes such as truck, bus, and train benefit from HRDA. This is also reflected in the visual examples in Fig.~\ref{fig:visual_examples}.

\begin{table}[tb]
\centering
\caption{Comparison with previous UDA methods. The results of HRDA are averaged over 3 random seeds. Further methods are shown in the supplement.}
\label{tab:sota}
\setlength{\tabcolsep}{1.5pt}
\resizebox{\textwidth}{!}{%
\begin{tabular}{l|ccccccccccccccccccc|c}
\toprule
 & Road & S.walk & Build. & Wall & Fence & Pole & Tr.Light & Sign & Veget. & Terrain & Sky & Person & Rider & Car & Truck & Bus & Train & M.bike & Bike & mIoU\\
\toprule
\multicolumn{21}{c}{GTA5 $\to$ Cityscapes} \\
\toprule
CBST~\cite{zou2018unsupervised} & 91.8 & 53.5 & 80.5 & 32.7 & 21.0 & 34.0 & 28.9 & 20.4 & 83.9 & 34.2 & 80.9 & 53.1 & 24.0 & 82.7 & 30.3 & 35.9 & 16.0 & 25.9 & 42.8 & 45.9\\
DACS~\cite{tranheden2021dacs} & 89.9 & 39.7 & 87.9 & 30.7 & 39.5 & 38.5 & 46.4 & 52.8 & 88.0 & 44.0 & 88.8 & 67.2 & 35.8 & 84.5 & 45.7 & 50.2 & 0.0 & 27.3 & 34.0 & 52.1\\
CorDA\cite{wang2021domain} & 94.7 & 63.1 & 87.6 & 30.7 & 40.6 & 40.2 & 47.8 & 51.6 & 87.6 & 47.0 & 89.7 & 66.7 & 35.9 & 90.2 & 48.9 & 57.5 & 0.0 & 39.8 & 56.0 & 56.6\\
BAPA\cite{liu2021bapa} & 94.4 & 61.0 & 88.0 & 26.8 & 39.9 & 38.3 & 46.1 & 55.3 & 87.8 & 46.1 & 89.4 & 68.8 & 40.0 & 90.2 & 60.4 & 59.0 & 0.0 & 45.1 & 54.2 & 57.4\\
ProDA~\cite{zhang2021prototypical} & 87.8 & 56.0 & 79.7 & 46.3 & 44.8 & 45.6 & 53.5 & 53.5 & 88.6 & 45.2 & 82.1 & 70.7 & 39.2 & 88.8 & 45.5 & 59.4 & 1.0 & 48.9 & 56.4 & 57.5\\
DAFormer~\cite{hoyer2021daformer} & \underline{95.7} & \underline{70.2} & \underline{89.4} & \underline{53.5} & \underline{48.1} & \underline{49.6} & \underline{55.8} & \underline{59.4} & \underline{89.9} & \underline{47.9} & \underline{92.5} & \underline{72.2} & \underline{44.7} & \underline{92.3} & \underline{74.5} & \underline{78.2} & \underline{65.1} & \underline{55.9} & \underline{61.8} & \underline{68.3}\\
HRDA (Ours) & \textbf{96.4} & \textbf{74.4} & \textbf{91.0} & \textbf{61.6} & \textbf{51.5} & \textbf{57.1} & \textbf{63.9} & \textbf{69.3} & \textbf{91.3} & \textbf{48.4} & \textbf{94.2} & \textbf{79.0} & \textbf{52.9} & \textbf{93.9} & \textbf{84.1} & \textbf{85.7} & \textbf{75.9} & \textbf{63.9} & \textbf{67.5} & \textbf{73.8}\\

\toprule
\multicolumn{21}{c}{Synthia $\to$ Cityscapes} \\
\toprule

CBST~\cite{zou2018unsupervised} & 68.0 & 29.9 & 76.3 & 10.8 & 1.4 & 33.9 & 22.8 & 29.5 & 77.6 & -- & 78.3 & 60.6 & 28.3 & 81.6 & -- & 23.5 & -- & 18.8 & 39.8 & 42.6\\
DACS~\cite{tranheden2021dacs} & 80.6 & 25.1 & 81.9 & 21.5 & 2.9 & 37.2 & 22.7 & 24.0 & 83.7 & -- & \underline{90.8} & 67.6 & 38.3 & 82.9 & -- & 38.9 & -- & 28.5 & 47.6 & 48.3\\
BAPA~\cite{liu2021bapa} & \underline{91.7} & \underline{53.8} & 83.9 & 22.4 & 0.8 & 34.9 & 30.5 & 42.8 & \underline{86.6} & -- & 88.2 & 66.0 & 34.1 & 86.6 & -- & 51.3 & -- & 29.4 & 50.5 & 53.3\\
CorDA~\cite{wang2021domain} & \textbf{93.3} & \textbf{61.6} & 85.3 & 19.6 & \underline{5.1} & 37.8 & 36.6 & 42.8 & 84.9 & -- & 90.4 & 69.7 & 41.8 & 85.6 & -- & 38.4 & -- & 32.6 & 53.9 & 55.0\\
ProDA~\cite{zhang2021prototypical} & 87.8 & 45.7 & 84.6 & 37.1 & 0.6 & 44.0 & 54.6 & 37.0 & \textbf{88.1} & -- & 84.4 & \underline{74.2} & 24.3 & \underline{88.2} & -- & 51.1 & -- & 40.5 & 45.6 & 55.5\\
DAFormer~\cite{hoyer2021daformer} & 84.5 & 40.7 & \underline{88.4} & \underline{41.5} & \textbf{6.5} & \underline{50.0} & \underline{55.0} & \underline{54.6} & 86.0 & -- & 89.8 & 73.2 & \underline{48.2} & 87.2 & -- & \underline{53.2} & -- & \underline{53.9} & \underline{61.7} & \underline{60.9}\\
HRDA (Ours) & 85.2 & 47.7 & \textbf{88.8} & \textbf{49.5} & 4.8 & \textbf{57.2} & \textbf{65.7} & \textbf{60.9} & 85.3 & -- & \textbf{92.9} & \textbf{79.4} & \textbf{52.8} & \textbf{89.0} & -- & \textbf{64.7} & -- & \textbf{63.9} & \textbf{64.9} & \textbf{65.8}\\
\bottomrule
\end{tabular}
}
\end{table}

\subsection{HRDA for Different UDA Methods}

HRDA is designed to be applicable to most UDA methods. In Tab.~\ref{tab:other_uda}, we compare the performance without and with HRDA of three further representative UDA methods. It can be seen that HRDA consistently improves the performance by at least +2.4 mIoU, demonstrating the importance of high- and multi-resolution inputs for UDA in general. Also, it shows that HRDA can be applied to different network architectures. The highest improvement is achieved for self-training methods (row 3-5) with +5.5 mIoU and more, which shows that the HRDA pseudo-labels positively reinforce the UDA process.

\begin{table}[tb]
\centering
\caption{HRDA applied to different UDA methods on GTA$\rightarrow$Cityscapes. Mean and standard deviation are provided over 3 random seeds.}
\label{tab:other_uda}
\setlength{\tabcolsep}{5pt}
\scriptsize
\begin{tabular}{rlllll}
\toprule
  &                          UDA Method &                           Network &       w/o HRDA &        w/ HRDA & Improvement \\
\midrule
1 &    Entropy Min. \cite{vu2019advent} &  DeepLabV2 \cite{chen2017deeplab} & 44.3\spm{0.4} & 46.7\spm{1.2} &        +2.4 \\
2 & Adversarial \cite{tsai2018learning} &  DeepLabV2 \cite{chen2017deeplab} & 44.2\spm{0.1} & 47.1\spm{1.0} &        +2.9 \\
3 &       DACS \cite{tranheden2021dacs} &  DeepLabV2 \cite{chen2017deeplab} & 53.9\spm{0.6} & 59.4\spm{1.2} &        +5.5 \\
4 &   DAFormer \cite{hoyer2021daformer} & DeepLabV2 \cite{chen2017deeplab} & 56.0\spm{0.5}  & 63.0\spm{0.4} & +7.0     \\
5 &   DAFormer \cite{hoyer2021daformer} & DAFormer \cite{hoyer2021daformer} & 68.3\spm{0.5} & 73.8\spm{0.3} &        +5.5 \\
\bottomrule
\end{tabular}
\end{table}

\subsection{Influence of Resolution and Crop Size on UDA}
\label{sec:exp_resolution_crop_size}

\begin{figure}[tb]
\centering
\includegraphics[width=\linewidth]{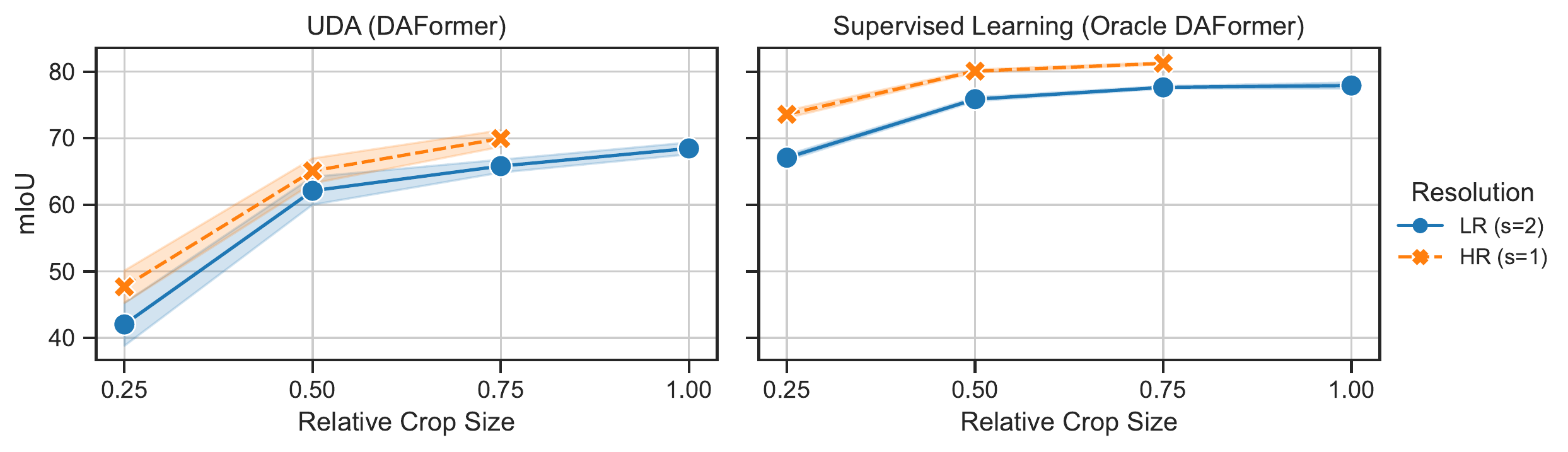}
\caption{Segmentation performance over the relative crop size ($h / \frac{H_T}{s}$) for different resolutions and for both the UDA method DAFormer~\cite{hoyer2021daformer} and the target-supervised oracle. There is no value for $\text{HR}_{1.0}$ due to GPU memory constraints.}
\label{fig:crop_resolution}
\end{figure}

\begin{figure}[tb]
\centering
\includegraphics[width=\linewidth]{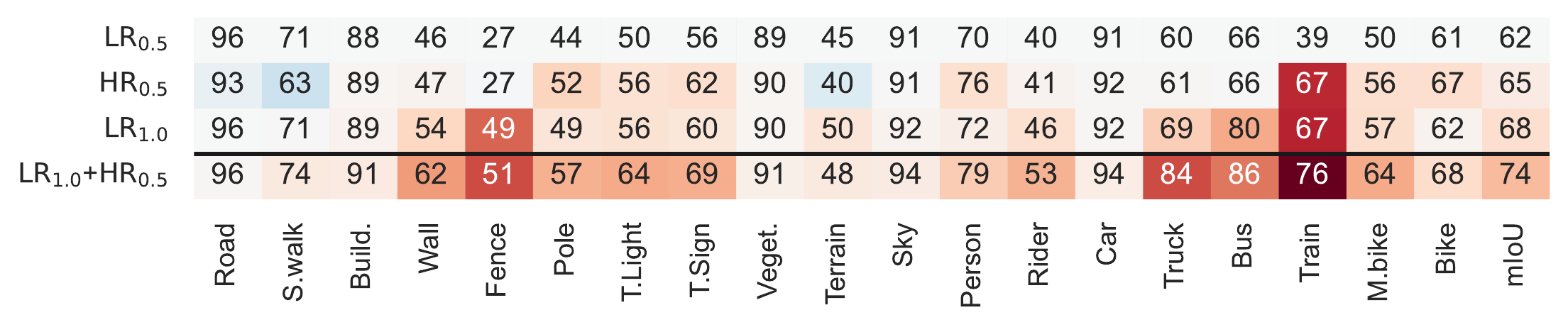}
\caption{
Class-wise IoU for the UDA method DAFormer~\cite{hoyer2021daformer} for different crop resolutions $\mathit{XR}$ ($s_\mathit{LR}{=}2, s_\mathit{HR}{=}1$) and relative crop sizes $a{=}h / \frac{H_T}{s_\mathit{XR}}$. Crops are denoted as $\mathit{XR}_a$. The colors indicate the difference to the first row.
}
\label{fig:crop_resolution_heatmap}
\end{figure}

In the following, we analyze the underlying principles of HRDA on GTA$\rightarrow$ Cityscapes, starting with the influence of the resolution and crop size on UDA.
For the comparison, we use the relative crop size $h / \frac{H_T}{s}$, which is normalized by the image height at the corresponding resolution, to disentangle the crop size from the used image resolution.
Fig.~\ref{fig:crop_resolution} shows that both an increased resolution and crop size improve the performance for both UDA and supervised learning. 
A large crop size is even more important for UDA than for supervised learning, i.e. a 4 times smaller LR crop reduces the performance by 39\% for UDA and by 14\% for supervised training. The larger crop provides more context clues and improves the performance of all classes, especially the ones that are difficult to adapt such as wall, fence, truck, bus, and train (cf. row 1 and 3 in Fig.~\ref{fig:crop_resolution_heatmap}), probably, as the relevant context clues are more domain-invariant~\cite{huang2020contextual, yang2021context, zhou2021context}.
A higher input resolution improves the UDA performance by a similar amount as it improves supervised learning. The improvement originates from a higher IoU for small classes such as pole, traffic light, traffic sign, person, motorbike, and bicycle, while some large classes such as road, sidewalk, and terrain have a decreased performance (cf. row 1 and 2 in Fig.~\ref{fig:crop_resolution_heatmap}). This supports that large objects are easier to adapt at LR while small objects are easier to adapt at HR, which can be exploited by the multi-resolution fusion of HRDA.


\subsection{Combining Crops from Multiple Resolutions with HRDA}
\label{sec:exp_multi_resolution_uda}
\noindent\textbf{Multi-Resolution UDA:} Next, we combine crops from LR and HR using the proposed multi-resolution training for UDA. Tab.~\ref{tab:context_crop_size} shows that training with multiple resolutions improves the performance over both LR-only and HR-only training by +3.4 mIoU (cf. row 2 and 3), which demonstrates that multi-resolution fusion with scale attention results in better domain adaptation. 

\noindent\textbf{Context Crop Size:} Based on the observation that large crops are important for UDA (Sec.~\ref{sec:exp_resolution_crop_size}), we increase the context crop size while keeping the detail crop size fixed, which further improves the performance by +5.3 mIoU (cf. row 3 and 5), demonstrating the effectiveness of the proposed small HR detail and large LR context crops. Fig.~\ref{fig:crop_resolution_heatmap} shows that the multi-resolution training combines the strength of the single-scale training with $\text{HR}_{0.5}$ and $\text{LR}_{1.0}$ as the multi-resolution IoU of each class is better than the best single-scale IoU (cf. row 2, 3, and 4).

\begin{table*}[tb]
\parbox[t]{.48\linewidth}{\centering
\caption{
HRDA context size. $\mathit{XR}_a$ denotes crops with resolution $\mathit{
XR}$ ($s_\mathit{LR}{=}2$, $s_\mathit{HR}{=}1$) and relative crop size $a{=}h / \frac{H_T}{s_\mathit{XR}}$.
}
\label{tab:context_crop_size}
\setlength{\tabcolsep}{5pt}
\scriptsize
\begin{tabular}{llll}
\toprule
  & Context Crop & Detail Crop &           mIoU \\
\midrule
 1 &  $\text{LR}_{0.5}$ &       -- & 62.1\spm{2.1} \\
 2 &      -- &   $\text{HR}_{0.5}$ & 65.1\spm{1.9} \\
 3 &  $\text{LR}_{0.5}$ &   $\text{HR}_{0.5}$ & 68.5\spm{0.6} \\
 4 & $\text{LR}_{0.75}$ &   $\text{HR}_{0.5}$ & 71.1\spm{1.7} \\
 5 &  $\text{LR}_{1.0}$ &   $\text{HR}_{0.5}$ & 73.8\spm{0.3} \\
\bottomrule
\end{tabular}
}
\hfill
\parbox[t]{.48\linewidth}{
\centering
\caption{
HRDA detail size. $\mathit{XR}_a$ denotes crops with resolution $\mathit{
XR}$ ($s_\mathit{LR}{=}2$, $s_\mathit{HR}{=}1$) and relative crop size $a{=}h / \frac{H_T}{s_\mathit{XR}}$.
}
\label{tab:detail_crop_size}
\setlength{\tabcolsep}{5pt}
\scriptsize
\begin{tabular}{llll}
\toprule
  & Context Crop & Detail Crop &           mIoU \\
\midrule
1 &  $\text{LR}_{1.0}$ &       -- & 68.5\spm{0.9} \\
2 &      -- &  $\text{HR}_{0.25}$ & 47.7\spm{2.4} \\
3 &  $\text{LR}_{1.0}$ &  $\text{HR}_{0.25}$ & 70.6\spm{0.7} \\
4 &  $\text{LR}_{1.0}$ & $\text{HR}_{0.375}$ & 71.7\spm{0.4} \\
5 &  $\text{LR}_{1.0}$ &   $\text{HR}_{0.5}$ & 73.8\spm{0.3} \\
\bottomrule
\end{tabular}
}
\end{table*}

\begin{table*}[tb]
\parbox[t]{.48\linewidth}{
\centering
\caption{Comparison of HRDA with naive HR crops that have a comparable GPU memory footprint ($\text{HR}_{0.75}$).}
\label{tab:baselines}
\setlength{\tabcolsep}{3pt}
\scriptsize
\begin{tabular}{lllll}
\toprule
{} & Context & Detail & Mem. &            mIoU \\
\midrule
1 &           -- &     $\text{HR}_{0.75}$ &  22.0 GB &  70.0\spm{1.2} \\
2 &      $\text{LR}_{0.75}$ &    $\text{HR}_{0.375}$ &  13.5 GB &  71.3\spm{0.3} \\
3 &       $\text{LR}_{1.0}$ &      $\text{HR}_{0.5}$ &  22.5 GB &  73.8\spm{0.3} \\
\bottomrule
\end{tabular}
}
\hfill
\parbox[t]{.48\linewidth}{
\centering
\caption{HRDA detail crop variants. Up-LR: LR crop upsampled to HR resolution.}
\vspace{0.15cm} 
\label{tab:detail_crop_ablations}
\setlength{\tabcolsep}{5pt}
\scriptsize
\begin{tabular}{llll}
\toprule
  & Context Crop & Detail Crop &           mIoU \\
\midrule
1 &       $\text{LR}_{1.0}$ &          -- & 68.5\spm{0.9} \\
2 &       $\text{LR}_{1.0}$ &      $\text{LR}_{0.5}$ & 69.1\spm{0.4} \\
3 &       $\text{LR}_{1.0}$ &   $\text{Up-LR}_{0.5}$ & 71.9\spm{1.5} \\
4 &       $\text{LR}_{1.0}$ &      $\text{HR}_{0.5}$ & 73.8\spm{0.3} \\
\bottomrule
\end{tabular}

}
\end{table*}

\noindent\textbf{Detail Crop Size:} 
Already the combination of the context crop $\text{LR}_{1.0}$ with an even smaller detail crop $\text{HR}_{0.25}$ outperforms training with solely the context crop by $+2.1$ mIoU (cf. row 1 and 3 in Tab.~\ref{tab:detail_crop_size}), even though $\text{HR}_{0.25}$ alone performs $-20.8$ mIoU worse (cf. row 1 and 2). This shows that the multi-resolution fusion effectively exploits the strength of the small detail crop while compensating for its lacking long-range dependencies with the context crop.
Further increasing the detail crop size, results in additional performance gains (cf. row 2 and 5). This shows that even though context information is not crucial for the detail crop, it is still helpful to some extent while being limited due to GPU memory constraints.

\noindent\textbf{Detail Crop Variants:} It is crucial for HRDA to use context/detail crops with different resolutions. Using LR instead of HR for the detail crop gives only a marginal gain of +0.6 over the baseline (cf. row 1 and 2 in Tab.~\ref{tab:detail_crop_ablations}). However, using an LR crop that is bilinearly upsampled to HR as detail crop does improve the performance by +3.4 mIoU (cf. row 1 and 3 in Tab.~\ref{tab:detail_crop_ablations}) but is still -1.9 mIoU worse than using a real HR detail crop (cf. row 3 and 4 in Tab.~\ref{tab:detail_crop_ablations}). This shows that the improved performance of HRDA comes from both the additional zoomed-in context information as well as the additional details in the HR image.

\noindent\textbf{Comparison with Naive HR:} We compare HRDA with naive large HR crops ($\text{HR}_{0.75}$), which have a comparable GPU memory footprint as HRDA. This is a very strong baseline, which is already +1.7 mIoU better than DAFormer \cite{hoyer2021daformer}. Tab.~\ref{tab:baselines} shows that HRDA still outperforms $\text{HR}_{0.75}$ crops by +3.8 mIoU (cf. row 1 and 3). Even when reducing the crop size of HRDA to match the size of $\text{HR}_{0.75}$, HRDA is still +1.3 mIoU better while requiring 40\% less GPU memory. This demonstrates that combining LR context crop and HR detail crop performs better than naively increasing the resolution, due to HRDA's capability of capturing large context information and multi-resolution fusion.

\noindent\textbf{HRDA Component Ablations:}
The components of HRDA are ablated in Tab.~\ref{tab:ablations}.
The most crucial component is the learned scale attention. While naively averaging the predictions from both scales gives no improvement over just using the context crop (cf. row 2 and 3), the learned scale attention improves the performance by +3.0 mIoU (cf. row 2 and 4). This shows that it is crucial to learn which scale is best-suited to adapt certain image regions. Generating pseudo-labels with different context views by overlapping slide detail crops results in a further gain of +0.9 mIoU (cf. row 4 and 5). Finally, additional supervision of the detail crop ($\lambda_d=0.1$) further provides +1.4 mIoU (cf. row 5 and 6).

\begin{table}[tb]
\centering
\caption{Component ablation of HRDA.}
\label{tab:ablations}
\setlength{\tabcolsep}{5pt}
\scriptsize
\begin{tabular}{lllllll}
\toprule
  & Context & Detail & Scale Attention & Overlapping Detail & Detail Loss &           mIoU \\
\midrule
1 &      -- &    \cm &              -- &                 -- &          -- & 65.1 \spm{1.9} \\
2 &     \cm &     -- &              -- &                 -- &          -- & 68.5 \spm{0.9} \\
3 &     \cm &    \cm &         Average &                 -- &          -- & 67.5 \spm{0.8} \\
4 &     \cm &    \cm &         Learned &                 -- &          -- & 71.5 \spm{0.5} \\
5 &     \cm &    \cm &         Learned &                \cm &          -- & 72.4 \spm{0.1} \\
6 &     \cm &    \cm &         Learned &                \cm &         \cm & 73.8 \spm{0.3} \\
\bottomrule
\end{tabular}
\end{table}

\begin{figure*}[tb]
\centering
\input{att/attention_head}
\includegraphics[width=\linewidth]{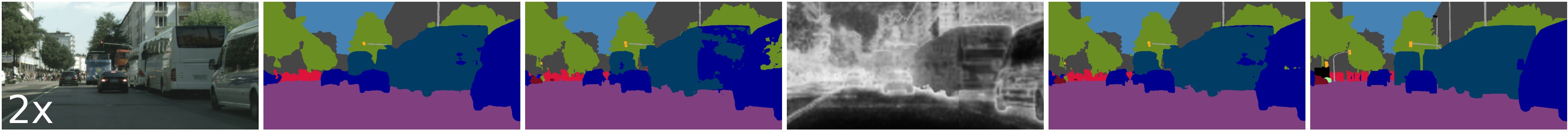}
\includegraphics[width=\linewidth]{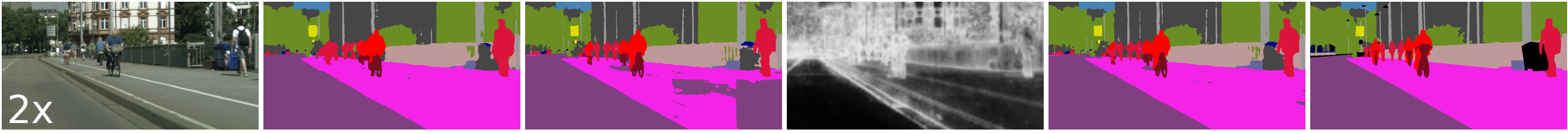}
\includegraphics[width=\linewidth]{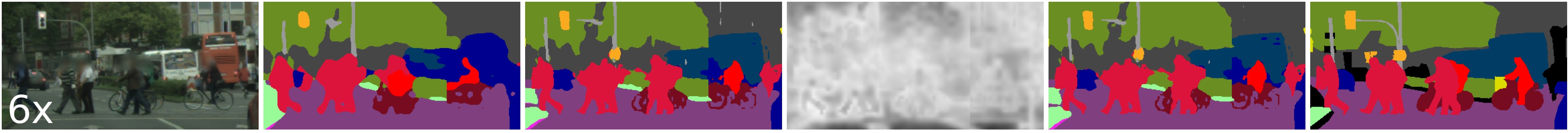}
\caption{Visual examples of the different predictions and the scale attention of HRDA. Large objects are better segmented from LR while small objects are better segmented from HR. The scale attention learns to utilize this pattern for fusing LR and HR predictions. The examples are zoomed in (2x or 6x) for better visibility of the details.}
\label{fig:attention_examples}
\end{figure*}

\noindent\textbf{Qualitative Analysis:}
Fig.~\ref{fig:attention_examples} provides representative visual examples demonstrating that LR predictions work better for large objects such as a bus or sidewalk (row 1/2) while HR predictions work better for small objects and fine details (row 3). The scale attention focuses on LR for large objects and on HR for small objects and segmentation borders, combining the strength of both.

\noindent\textbf{Supplement:} The supplement provides further parameter studies, additional baseline comparisons, a runtime analysis, and an extended qualitative analysis.

%% file: preds/prediction_head.tex
{\footnotesize
\begin{tabularx}{\linewidth}{*{5}{Y}}
Image & ProDA~\cite{zhang2021prototypical} & DAFormer~\cite{hoyer2021daformer} & HRDA (Ours) & Ground Truth \\
\end{tabularx}
} %

%% file: att/attention_head.tex
{\footnotesize
\begin{tabularx}{\linewidth}{*{6}{Y}}
Image & LR Pred. & HR Pred. & Scale Attent. & Fused Pred. & G. Truth \\
\end{tabularx}
} %

%% file: tex_conclusions.tex
\section{Conclusions}

In this work, we presented HRDA, a multi-resolution approach for UDA that combines the advantages of small HR detail crops and large LR context crops using a learned scale attention, while maintaining a manageable GPU memory footprint. It can be combined with various UDA methods and achieves a consistent, significant improvement. Overall, HRDA achieves an unprecedented performance of 73.8 mIoU on GTA$\rightarrow$Cityscapes and 65.8 mIoU on Synthia$\rightarrow$Cityscapes, which is a respective gain of +5.5 mIoU and +4.9 mIoU over the previous SOTA.

\noindent\textbf{Acknowledgements:} This work is supported by the European Lighthouse on Secure and Safe AI (ELSA) and a Facebook Academic Gift on Robust Perception (INFO224).

%% file: tex_supplement.tex
\section{Overview}

In the supplementary material for HRDA, we provide the source code (Sec.~\ref{sec:supp_further_implementation_details}), additional experimental analysis (Sec.~\ref{sec:supp_detail_loss_weight} and \ref{sec:supp_context_scale}), comparisons with further baselines (Sec.~\ref{sec:further_baselines}), an analysis of the runtime (Sec.~\ref{sec:runtime}), an extended comparison with previous UDA methods (Sec.~\ref{sec:supp_sota}), and a comprehensive qualitative analysis of the predictions from HRDA (Sec.~\ref{sec:supp_examples}).

\section{Source Code}
\label{sec:supp_further_implementation_details}

The source code to reproduce HRDA and all ablation
studies is provided at \url{https://github.com/lhoyer/HRDA}. Please, refer to the contained \texttt{README.md} for further instructions to set up the environment and run the experiments. Our implementation is based on the DAFormer framework~\cite{hoyer2021daformer} and the mmsegmentation framework~\cite{mmseg2020}.

\section{Influence of Detail Loss Weight}
\label{sec:supp_detail_loss_weight}

\begin{figure}
\centering
\includegraphics[width=0.8\linewidth]{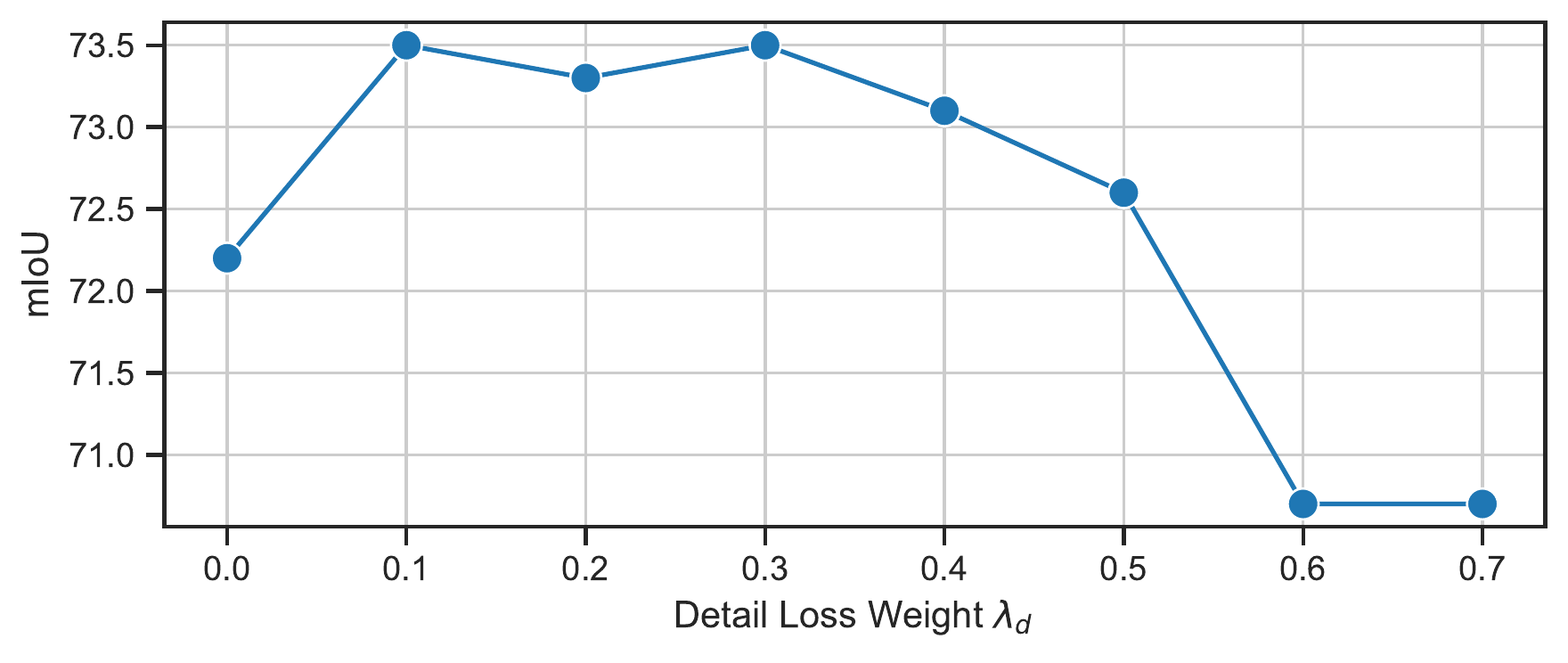}
\caption{Study of the UDA performance of HRDA with respect to the detail loss weight $\lambda_d$ on GTA$\rightarrow$Cityscapes.}
\label{fig:detail_loss_weight}
\end{figure}

In Fig.~\ref{fig:detail_loss_weight}, the sensitivity of the UDA performance of HRDA with respect to the detail loss weight $\lambda_d$ is studied. It is shown that values in the range between 0.1 and 0.3 give a consistently good UDA performance, which is a reasonably broad range for a robust hyperparameter choice. If $\lambda_d$ is either too small or too large, HRDA focuses too much on LR or HR, respectively.

\section{Influence of Context Scale}
\label{sec:supp_context_scale}

\begin{table}
\centering
\caption{Influence of the context downscale factor $s$ on HRDA performance. A larger downscale factor $s$ results in a lower crop resolution. The relative crop size is $a{=}h / \frac{H_T}{s}$.}
\label{tab:context_scale}
\setlength{\tabcolsep}{5pt}
\scriptsize
\begin{tabular}{llllll}
\toprule
  & Context $s_c$ & Context Rel. Size $a_c$ & Detail $s_d$ & Detail Rel. Size $a_d$ &          mIoU \\
\midrule
1 &            -- &                      -- &       1 (HR) &                    0.5 & 65.1\spm{1.9} \\
2 &             4 &                     0.5 &       1 (HR) &                    0.5 & 65.9\spm{1.2} \\
3 &        2 (LR) &                     0.5 &       1 (HR) &                    0.5 & 68.5\spm{0.6} \\
4 &          1.33 &                     0.5 &       1 (HR) &                    0.5 & 67.5\spm{0.7} \\
\bottomrule
\end{tabular}
\end{table}

We further study the influence of the downscale factor of the context crop  $s_c$ in Tab.~\ref{tab:context_scale}. It can be seen that the default downscale factor $s_c=2$ provides the best performance. A context crop with a higher downscale factor $s_c=4$ performs worse by -2.6 mIoU than the default choice (cf. row 2 and 3) but is still slightly better than just using the detail crop by +0.8 mIoU (cf. row 1 and 2). We assume that with $s_c=4$ the resolution of the context crop is too low to be useful for UDA. A context crop with a small downscale factor $s_c=1.33$ performs better than the high downscale factor $s_c=4$ by +1.6 mIoU (cf. row 2 and 4) but still does not achieve the performance of the default $s_c=2$ with a difference of -1.0 mIoU. Possibly, the resolution of $s_c=1.33$ is too similar to the detail crop resolution $s_d=1$ and, therefore, it does not provide a sufficiently different perspective on the data, which is important for multi-resolution UDA.

\section{Further Baselines}
\label{sec:further_baselines}

\subsection{Overlapping Sliding Window Inference (OSW)}

Prior works in the field of UDA (including DAFormer) use whole image inference while HRDA utilizes overlapping sliding window inference (OSW). To show that the improvement of HRDA is not mainly caused by the OSW inference, we also evaluate prior arts with OSW (see Tab.~\ref{tab:osw_hr} col. 4). It can be seen that OSW only slightly benefits DAFormer by +0.3 mIoU. Still, HRDA outperforms DAFormer with OSW by +5.2 mIoU.

\subsection{Naive High-Resolution UDA}

In Sec. 5.5 of the main paper, we compared HRDA with the naive HR crops ($\text{HR}_{0.75}$) for DAFormer~\cite{hoyer2021daformer}. Here, we extend this comparison also to DACS~\cite{tranheden2021dacs}, which uses another UDA method and network architecture. Tab.~\ref{tab:osw_hr} col. 5 shows that naive HR training improves the performance of DACS similarly to DAFormer. HRDA outperforms naive HR training by +3.6 mIoU for DACS and +3.8 mIoU for DAFormer.

\begin{table}
\centering
\scriptsize
\setlength{\tabcolsep}{5pt}
\caption{Overlapping sliding window inference (OSW) and naive HR training for different UDA methods and network architectures on GTA$\rightarrow$Cityscapes.}
\label{tab:osw_hr}
\begin{tabular}{llllll}
\toprule
UDA Method     & Network & Baseline & OSW & Naive HR+OSW & HRDA \\
\midrule
DACS~\cite{tranheden2021dacs}      & DeepLabV2~\cite{chen2017deeplab} & 53.9\spm{0.6}     & 53.9\spm{0.6} & 55.8\spm{1.6}         & 59.4\spm{1.2}      \\
DAFormer~\cite{hoyer2021daformer}  & DAFormer~\cite{hoyer2021daformer} & 68.3\spm{0.5}     & 68.6\spm{0.3} & 70.0\spm{1.2}         & 73.8\spm{0.3}      \\
\bottomrule
\end{tabular}
\end{table}

\subsection{Scale-Invariance Loss}

As another baseline for HRDA, we integrate the scale-invariance loss of Guan et al.~\cite{guan2021scale} into DAFormer~\cite{hoyer2021daformer}. The optimal loss weight when combined with DAFormer is determined with a grid search as 0.1.
As shown in Tab.~\ref{tab:scale_consistency_baseline}, DAFormer with scale-invariance loss~\cite{guan2021scale} achieves 68.8 mIoU on GTA$\rightarrow$Cityscapes, which is only a small gain of +0.5 mIoU over DAFormer, while HRDA achieves +5.5 mIoU. We assume that the effect of the scale-invariance loss is not so pronounced as DAFormer is much stronger (68.3 mIoU) than the original baseline (43.8 mIoU).
This further emphasizes that the contribution of HRDA goes beyond scale consistency training.

\begin{table}
\centering
\caption{Comparison of scale consistency training and HRDA on GTA$\rightarrow$Cityscapes.}
\label{tab:scale_consistency_baseline}
\setlength{\tabcolsep}{5pt}
\scriptsize
\begin{tabular}{llll}
\toprule
 & Baseline & w/ Scale-Invariance~\cite{guan2021scale} & w/ HRDA \\
\midrule
Original~\cite{guan2021scale} & 43.8 & 48.1 & -- \\
DAFormer~\cite{hoyer2021daformer} & 63.3 & 63.8 & 73.8 \\
\bottomrule
\end{tabular}
\end{table}

\section{Training and Inference Time}
\label{sec:runtime}

The training of HRDA takes 32h on a Titan RTX. To put this into context, the training of other UDA methods~\cite{yang2020fda, zhang2021prototypical, liu2021bapa} can take several days. The runtime and memory consumption of HRDA during inference is shown in Tab.~\ref{tab:runtime}.
The inference runs with 0.8 img/s, 3.3 TFLOPs, and 9.4 GB GPU memory.
The focus of HRDA is UDA performance and not fast inference as the latter is not an inherent constraint of UDA.
Still, if efficient inference is important, \textit{non}-overlapping sliding window inference (stride equal to window size) can be used at test-time resulting in 2.2 img/s, 1.8 TFLOPs, and 3.2 GB with still 73.4 mIoU.
Alternatively, the knowledge of HRDA can be distilled into a faster single-scale DeepLabV2~\cite{chen2017deeplab} model, which is commonly used for UDA. For that purpose, the multi-resolution HRDA is utilized to generate high-quality pseudo-labels for the target domain and the single-scale DeepLabV2 model is trained on the target domain with a pixel-wise cross-entropy loss using the pseudo-labels.
The distilled DeepLabV2 model achieves 70.4 mIoU at 3.4 img/s, 1.4 TFLOPs, and 1.3 GB. Both results are significantly better than the previous SOTA performance of DAFormer~\cite{hoyer2021daformer}, which is 68.3 mIoU.

\begin{table}
\centering
\scriptsize
\setlength{\tabcolsep}{4pt}
\caption{Runtime and memory consumption of HRDA variants during inference on an Nvidia Titan RTX.}
\label{tab:runtime}
\begin{tabular}{lllll}
\toprule
                           & Throughput (img/s) & TFLOPs & GPU Mem. (GB) & mIoU \\
\midrule
HRDA                       & 0.8                & 3.3    & 9.4              & 73.8 \\
HRDA w/ Non-Overlapping SW & 2.2                & 1.8    & 3.2              & 73.4 \\
HRDA w/ Distilled DeepLabV2& 3.4                & 1.4    & 1.3              & 70.4 \\
\bottomrule
\end{tabular}
\end{table}

\section{Extended Comparison with Previous UDA Methods}
\label{sec:supp_sota}

We extend the comparison of HRDA with previous UDA methods from the main paper by a large selection of further methods for GTA$\rightarrow$Cityscapes in Tab.~\ref{tab:sota_gta} and for Synthia$\rightarrow$Cityscapes in Tab.~\ref{tab:sota_synthia}. It can be seen that HRDA$_\text{DAFormer}$ (the default HRDA based on DAFormer) still outperforms previous UDA methods by a large margin both for the class-wise IoU as well as the overall mIoU. The highest performance gains are achieved for classes with fine segmentation details such as pole, traffic light, traffic sign, person, rider, motorbike, and bike. But also large classes such as truck, bus, and train benefit from HRDA.
Only a few classes such as road, sidewalk, fence, and vegetation on Synthia$\rightarrow$Cityscapes have a lower performance than the respective best comparison method. 
The comparably low performance is probably inherited from DAFormer~\cite{hoyer2021daformer}, which is the basis of HRDA. 
This effect might be caused by the shape bias of the used Transformer encoder as discussed in DAFormer~\cite{hoyer2021daformer}.
Possibly, the performance for the mentioned stuff classes could be improved for HRDA by integrating the depth-clues as done in CorDA~\cite{wang2021domain} or pseudo-label prototypes as used in ProDA~\cite{zhang2021prototypical}. Further,  Tab.~\ref{tab:sota_gta} and Tab.~\ref{tab:sota_synthia} show the performance of HRDA when used with a DeepLabV2 network instead of a DAFormer network. It can be seen that HRDA$_\text{DeepLabV2}$ outperforms all DeepLabV2-based UDA methods (all methods except DAFormer) on GTA$\rightarrow$Cityscapes and that it even outperforms DAFormer on Synthia$\rightarrow$Cityscapes.

\begin{table}[htb]
\centering
\caption{Comparison with previous UDA methods on GTA$\rightarrow$Cityscapes. The results of HRDA are averaged over 3 random seeds.}
\label{tab:sota_gta}
\setlength{\tabcolsep}{1.5pt}
\resizebox{\textwidth}{!}{%
\begin{tabular}{l|ccccccccccccccccccc|c}
\toprule
 & Road & S.walk & Build. & Wall & Fence & Pole & Tr.Light & Sign & Veget. & Terrain & Sky & Person & Rider & Car & Truck & Bus & Train & M.bike & Bike & mIoU\\
\midrule
AdaptSeg~\cite{tsai2018learning} & 86.5 & 25.9 & 79.8 & 22.1 & 20.0 & 23.6 & 33.1 & 21.8 & 81.8 & 25.9 & 75.9 & 57.3 & 26.2 & 76.3 & 29.8 & 32.1 & 7.2 & 29.5 & 32.5 & 41.4\\
CyCADA~\cite{hoffman2018cycada} & 86.7 & 35.6 & 80.1 & 19.8 & 17.5 & 38.0 & 39.9 & 41.5 & 82.7 & 27.9 & 73.6 & 64.9 & 19.0 & 65.0 & 12.0 & 28.6 & 4.5 & 31.1 & 42.0 & 42.7\\
CLAN~\cite{luo2019taking} & 87.0 & 27.1 & 79.6 & 27.3 & 23.3 & 28.3 & 35.5 & 24.2 & 83.6 & 27.4 & 74.2 & 58.6 & 28.0 & 76.2 & 33.1 & 36.7 & 6.7 & 31.9 & 31.4 & 43.2\\
ADVENT~\cite{vu2019advent} & 89.4 & 33.1 & 81.0 & 26.6 & 26.8 & 27.2 & 33.5 & 24.7 & 83.9 & 36.7 & 78.8 & 58.7 & 30.5 & 84.8 & 38.5 & 44.5 & 1.7 & 31.6 & 32.4 & 45.5\\
APODA~\cite{yang2020adversarial} & 85.6 & 32.8 & 79.0 & 29.5 & 25.5 & 26.8 & 34.6 & 19.9 & 83.7 & 40.6 & 77.9 & 59.2 & 28.3 & 84.6 & 34.6 & 49.2 & 8.0 & 32.6 & 39.6 & 45.9\\
CBST~\cite{zou2018unsupervised} & 91.8 & 53.5 & 80.5 & 32.7 & 21.0 & 34.0 & 28.9 & 20.4 & 83.9 & 34.2 & 80.9 & 53.1 & 24.0 & 82.7 & 30.3 & 35.9 & 16.0 & 25.9 & 42.8 & 45.9\\
PatchAlign~\cite{tsai2019domain} & 92.3 & 51.9 & 82.1 & 29.2 & 25.1 & 24.5 & 33.8 & 33.0 & 82.4 & 32.8 & 82.2 & 58.6 & 27.2 & 84.3 & 33.4 & 46.3 & 2.2 & 29.5 & 32.3 & 46.5\\
MRKLD~\cite{zou2019confidence} & 91.0 & 55.4 & 80.0 & 33.7 & 21.4 & 37.3 & 32.9 & 24.5 & 85.0 & 34.1 & 80.8 & 57.7 & 24.6 & 84.1 & 27.8 & 30.1 & 26.9 & 26.0 & 42.3 & 47.1\\
BDL~\cite{li2019bidirectional} & 91.0 & 44.7 & 84.2 & 34.6 & 27.6 & 30.2 & 36.0 & 36.0 & 85.0 & 43.6 & 83.0 & 58.6 & 31.6 & 83.3 & 35.3 & 49.7 & 3.3 & 28.8 & 35.6 & 48.5\\
FADA~\cite{wang2020classes} & 91.0 & 50.6 & 86.0 & 43.4 & 29.8 & 36.8 & 43.4 & 25.0 & 86.8 & 38.3 & 87.4 & 64.0 & 38.0 & 85.2 & 31.6 & 46.1 & 6.5 & 25.4 & 37.1 & 50.1\\
CAG~\cite{zhang2019category} & 90.4 & 51.6 & 83.8 & 34.2 & 27.8 & 38.4 & 25.3 & 48.4 & 85.4 & 38.2 & 78.1 & 58.6 & 34.6 & 84.7 & 21.9 & 42.7 & 41.1 & 29.3 & 37.2 & 50.2\\
Seg-Uncert.~\cite{zheng2021rectifying} & 90.4 & 31.2 & 85.1 & 36.9 & 25.6 & 37.5 & 48.8 & 48.5 & 85.3 & 34.8 & 81.1 & 64.4 & 36.8 & 86.3 & 34.9 & 52.2 & 1.7 & 29.0 & 44.6 & 50.3\\
FDA\cite{yang2020fda} & 92.5 & 53.3 & 82.4 & 26.5 & 27.6 & 36.4 & 40.6 & 38.9 & 82.3 & 39.8 & 78.0 & 62.6 & 34.4 & 84.9 & 34.1 & 53.1 & 16.9 & 27.7 & 46.4 & 50.5\\
PIT\cite{lv2020cross} & 87.5 & 43.4 & 78.8 & 31.2 & 30.2 & 36.3 & 39.9 & 42.0 & 79.2 & 37.1 & 79.3 & 65.4 & 37.5 & 83.2 & 46.0 & 45.6 & 25.7 & 23.5 & 49.9 & 50.6\\
IAST\cite{mei2020instance} & 93.8 & 57.8 & 85.1 & 39.5 & 26.7 & 26.2 & 43.1 & 34.7 & 84.9 & 32.9 & 88.0 & 62.6 & 29.0 & 87.3 & 39.2 & 49.6 & 23.2 & 34.7 & 39.6 & 51.5\\
DACS~\cite{tranheden2021dacs} & 89.9 & 39.7 & 87.9 & 30.7 & 39.5 & 38.5 & 46.4 & 52.8 & 88.0 & 44.0 & 88.8 & 67.2 & 35.8 & 84.5 & 45.7 & 50.2 & 0.0 & 27.3 & 34.0 & 52.1\\
SAC~\cite{araslanov2021self} & 90.4 & 53.9 & 86.6 & 42.4 & 27.3 & 45.1 & 48.5 & 42.7 & 87.4 & 40.1 & 86.1 & 67.5 & 29.7 & 88.5 & 49.1 & 54.6 & 9.8 & 26.6 & 45.3 & 53.8\\
CTF~\cite{ma2021coarse} & 92.5 & 58.3 & 86.5 & 27.4 & 28.8 & 38.1 & 46.7 & 42.5 & 85.4 & 38.4 & 91.8 & 66.4 & 37.0 & 87.8 & 40.7 & 52.4 & 44.6 & 41.7 & 59.0 & 56.1\\
CorDA\cite{wang2021domain} & 94.7 & 63.1 & 87.6 & 30.7 & 40.6 & 40.2 & 47.8 & 51.6 & 87.6 & 47.0 & 89.7 & 66.7 & 35.9 & 90.2 & 48.9 & 57.5 & 0.0 & 39.8 & 56.0 & 56.6\\
BAPA\cite{liu2021bapa} & 94.4 & 61.0 & 88.0 & 26.8 & 39.9 & 38.3 & 46.1 & 55.3 & 87.8 & 46.1 & 89.4 & 68.8 & 40.0 & 90.2 & 60.4 & 59.0 & 0.0 & 45.1 & 54.2 & 57.4\\
ProDA~\cite{zhang2021prototypical} & 87.8 & 56.0 & 79.7 & 46.3 & 44.8 & 45.6 & 53.5 & 53.5 & 88.6 & 45.2 & 82.1 & 70.7 & 39.2 & 88.8 & 45.5 & 59.4 & 1.0 & 48.9 & 56.4 & 57.5\\
DAFormer~\cite{hoyer2021daformer} & 95.7 & 70.2 & 89.4 & \underline{53.5} & \underline{48.1} & \underline{49.6} & 55.8 & 59.4 & \underline{89.9} & \underline{47.9} & \underline{92.5} & 72.2 & 44.7 & \underline{92.3} & \underline{74.5} & \underline{78.2} & \underline{65.1} & \underline{55.9} & 61.8 & \underline{68.3}\\
\midrule
HRDA$_\text{DeepLabV2}$ & \underline{96.2} & \underline{73.1} & \underline{89.7} & 43.2 & 39.9 & 47.5 & \underline{60.0} & \underline{60.0} & \underline{89.9} & 47.1 & 90.2 & \underline{75.9} & \underline{49.0} & 91.8 & 61.9 & 59.3 & 10.2 & 47.0 & \underline{65.3} & 63.0\\
HRDA$_\text{DAFormer}$ & \textbf{96.4} & \textbf{74.4} & \textbf{91.0} & \textbf{61.6} & \textbf{51.5} & \textbf{57.1} & \textbf{63.9} & \textbf{69.3} & \textbf{91.3} & \textbf{48.4} & \textbf{94.2} & \textbf{79.0} & \textbf{52.9} & \textbf{93.9} & \textbf{84.1} & \textbf{85.7} & \textbf{75.9} & \textbf{63.9} & \textbf{67.5} & \textbf{73.8}\\
\bottomrule
\end{tabular}
}
\end{table}

\begin{table}[htb]
\centering
\caption{Comparison with previous UDA methods on Synthia$\rightarrow$Cityscapes. The results of HRDA are averaged over 3 random seeds.}
\label{tab:sota_synthia}
\setlength{\tabcolsep}{1.5pt}
\resizebox{\textwidth}{!}{%
\begin{tabular}{l|cccccccccccccccc|cc}
\toprule
 & Road & S.walk & Build. & Wall & Fence & Pole & Tr.Light & Sign & Veget. & Sky & Person & Rider & Car & Bus & M.bike & Bike & mIoU16 & mIoU13\\
\midrule
SPIGAN~\cite{lee2018spigan} & 71.1 & 29.8 & 71.4 & 3.7 & 0.3 & 33.2 & 6.4 & 15.6 & 81.2 & 78.9 & 52.7 & 13.1 & 75.9 & 25.5 & 10.0 & 20.5 & 36.8 & 42.4\\
GIO-Ada~\cite{chen2019learning} & 78.3 & 29.2 & 76.9 & 11.4 & 0.3 & 26.5 & 10.8 & 17.2 & 81.7 & 81.9 & 45.8 & 15.4 & 68.0 & 15.9 & 7.5 & 30.4 & 37.3 & 43.0\\
AdaptSeg~\cite{tsai2018learning} & 79.2 & 37.2 & 78.8 & -- & -- & -- & 9.9 & 10.5 & 78.2 & 80.5 & 53.5 & 19.6 & 67.0 & 29.5 & 21.6 & 31.3 & -- & 45.9\\
PatchAlign~\cite{tsai2019domain} & 82.4 & 38.0 & 78.6 & 8.7 & 0.6 & 26.0 & 3.9 & 11.1 & 75.5 & 84.6 & 53.5 & 21.6 & 71.4 & 32.6 & 19.3 & 31.7 & 40.0 & 46.5\\
CLAN~\cite{luo2019taking} & 81.3 & 37.0 & 80.1 & -- & -- & -- & 16.1 & 13.7 & 78.2 & 81.5 & 53.4 & 21.2 & 73.0 & 32.9 & 22.6 & 30.7 & -- & 47.8\\
ADVENT~\cite{vu2019advent} & 85.6 & 42.2 & 79.7 & 8.7 & 0.4 & 25.9 & 5.4 & 8.1 & 80.4 & 84.1 & 57.9 & 23.8 & 73.3 & 36.4 & 14.2 & 33.0 & 41.2 & 48.0\\
CBST~\cite{zou2018unsupervised} & 68.0 & 29.9 & 76.3 & 10.8 & 1.4 & 33.9 & 22.8 & 29.5 & 77.6 & 78.3 & 60.6 & 28.3 & 81.6 & 23.5 & 18.8 & 39.8 & 42.6 & 48.9\\
DADA~\cite{vu2019dada} & 89.2 & 44.8 & 81.4 & 6.8 & 0.3 & 26.2 & 8.6 & 11.1 & 81.8 & 84.0 & 54.7 & 19.3 & 79.7 & 40.7 & 14.0 & 38.8 & 42.6 & 49.8\\
MRKLD~\cite{zou2019confidence} & 67.7 & 32.2 & 73.9 & 10.7 & 1.6 & 37.4 & 22.2 & 31.2 & 80.8 & 80.5 & 60.8 & 29.1 & 82.8 & 25.0 & 19.4 & 45.3 & 43.8 & 50.1\\
BDL~\cite{li2019bidirectional} & 86.0 & 46.7 & 80.3 & -- & -- & -- & 14.1 & 11.6 & 79.2 & 81.3 & 54.1 & 27.9 & 73.7 & 42.2 & 25.7 & 45.3 & -- & 51.4\\
CAG~\cite{zhang2019category} & 84.7 & 40.8 & 81.7 & 7.8 & 0.0 & 35.1 & 13.3 & 22.7 & 84.5 & 77.6 & 64.2 & 27.8 & 80.9 & 19.7 & 22.7 & 48.3 & 44.5 & 51.5\\
PIT~\cite{lv2020cross} & 83.1 & 27.6 & 81.5 & 8.9 & 0.3 & 21.8 & 26.4 & 33.8 & 76.4 & 78.8 & 64.2 & 27.6 & 79.6 & 31.2 & 31.0 & 31.3 & 44.0 & 51.8\\
SIM~\cite{wang2020differential} & 83.0 & 44.0 & 80.3 & -- & -- & -- & 17.1 & 15.8 & 80.5 & 81.8 & 59.9 & 33.1 & 70.2 & 37.3 & 28.5 & 45.8 & -- & 52.1\\
FDA~\cite{yang2020fda} & 79.3 & 35.0 & 73.2 & -- & -- & -- & 19.9 & 24.0 & 61.7 & 82.6 & 61.4 & 31.1 & 83.9 & 40.8 & 38.4 & 51.1 & -- & 52.5\\
FADA~\cite{wang2020classes} & 84.5 & 40.1 & 83.1 & 4.8 & 0.0 & 34.3 & 20.1 & 27.2 & 84.8 & 84.0 & 53.5 & 22.6 & 85.4 & 43.7 & 26.8 & 27.8 & 45.2 & 52.5\\
APODA~\cite{yang2020adversarial} & 86.4 & 41.3 & 79.3 & -- & -- & -- & 22.6 & 17.3 & 80.3 & 81.6 & 56.9 & 21.0 & 84.1 & 49.1 & 24.6 & 45.7 & -- & 53.1\\
DACS~\cite{tranheden2021dacs} & 80.6 & 25.1 & 81.9 & 21.5 & 2.9 & 37.2 & 22.7 & 24.0 & 83.7 & \underline{90.8} & 67.6 & 38.3 & 82.9 & 38.9 & 28.5 & 47.6 & 48.3 & 54.8\\
Seg-Uncert.~\cite{zheng2021rectifying} & 87.6 & 41.9 & 83.1 & 14.7 & 1.7 & 36.2 & 31.3 & 19.9 & 81.6 & 80.6 & 63.0 & 21.8 & 86.2 & 40.7 & 23.6 & 53.1 & 47.9 & 54.9\\
CTF~\cite{ma2021coarse} & 75.7 & 30.0 & 81.9 & 11.5 & 2.5 & 35.3 & 18.0 & 32.7 & 86.2 & 90.1 & 65.1 & 33.2 & 83.3 & 36.5 & 35.3 & 54.3 & 48.2 & 55.5\\
IAST~\cite{mei2020instance} & 81.9 & 41.5 & 83.3 & 17.7 & 4.6 & 32.3 & 30.9 & 28.8 & 83.4 & 85.0 & 65.5 & 30.8 & 86.5 & 38.2 & 33.1 & 52.7 & 49.8 & 57.0\\
SAC~\cite{araslanov2021self} & 89.3 & 47.2 & 85.5 & 26.5 & 1.3 & 43.0 & 45.5 & 32.0 & 87.1 & 89.3 & 63.6 & 25.4 & 86.9 & 35.6 & 30.4 & 53.0 & 52.6 & 59.3\\
BAPA~\cite{liu2021bapa} & \underline{91.7} & \underline{53.8} & 83.9 & 22.4 & 0.8 & 34.9 & 30.5 & 42.8 & 86.6 & 88.2 & 66.0 & 34.1 & 86.6 & 51.3 & 29.4 & 50.5 & 53.3 & 61.2\\
ProDA~\cite{zhang2021prototypical} & 87.8 & 45.7 & 84.6 & 37.1 & 0.6 & 44.0 & 54.6 & 37.0 & \textbf{88.1} & 84.4 & 74.2 & 24.3 & \underline{88.2} & 51.1 & 40.5 & 45.6 & 55.5 & 62.0\\
CorDA~\cite{wang2021domain} & \textbf{93.3} & \textbf{61.6} & 85.3 & 19.6 & \underline{5.1} & 37.8 & 36.6 & 42.8 & 84.9 & 90.4 & 69.7 & 41.8 & 85.6 & 38.4 & 32.6 & 53.9 & 55.0 & 62.8\\
DAFormer~\cite{hoyer2021daformer} & 84.5 & 40.7 & \underline{88.4} & \underline{41.5} & \textbf{6.5} & 50.0 & 55.0 & 54.6 & 86.0 & 89.8 & 73.2 & 48.2 & 87.2 & \underline{53.2} & 53.9 & 61.7 & 60.9 & 67.4\\
\midrule
HRDA$_\text{DeepLabV2}$ & 85.8 & 47.3 & 87.3 & 27.3 & 1.4 & \underline{50.5} & \underline{57.8} & \textbf{61.0} & \underline{87.4} & 89.1 & \underline{76.2} & \underline{48.5} & 87.3 & 49.3 & \underline{55.0} & \textbf{68.2} & \underline{61.2} & \underline{69.2}\\
HRDA$_\text{DAFormer}$ & 85.2 & 47.7 & \textbf{88.8} & \textbf{49.5} & 4.8 & \textbf{57.2} & \textbf{65.7} & \underline{60.9} & 85.3 & \textbf{92.9} & \textbf{79.4} & \textbf{52.8} & \textbf{89.0} & \textbf{64.7} & \textbf{63.9} & \underline{64.9} & \textbf{65.8} & \textbf{72.4}\\
\bottomrule
\end{tabular}
}
\end{table}

\FloatBarrier

\section{Further Qualitative Examples}
\label{sec:supp_examples}

\begin{figure*}
\centering
\input{supp_preds/prediction_head}
\includegraphics[width=0.31\linewidth]{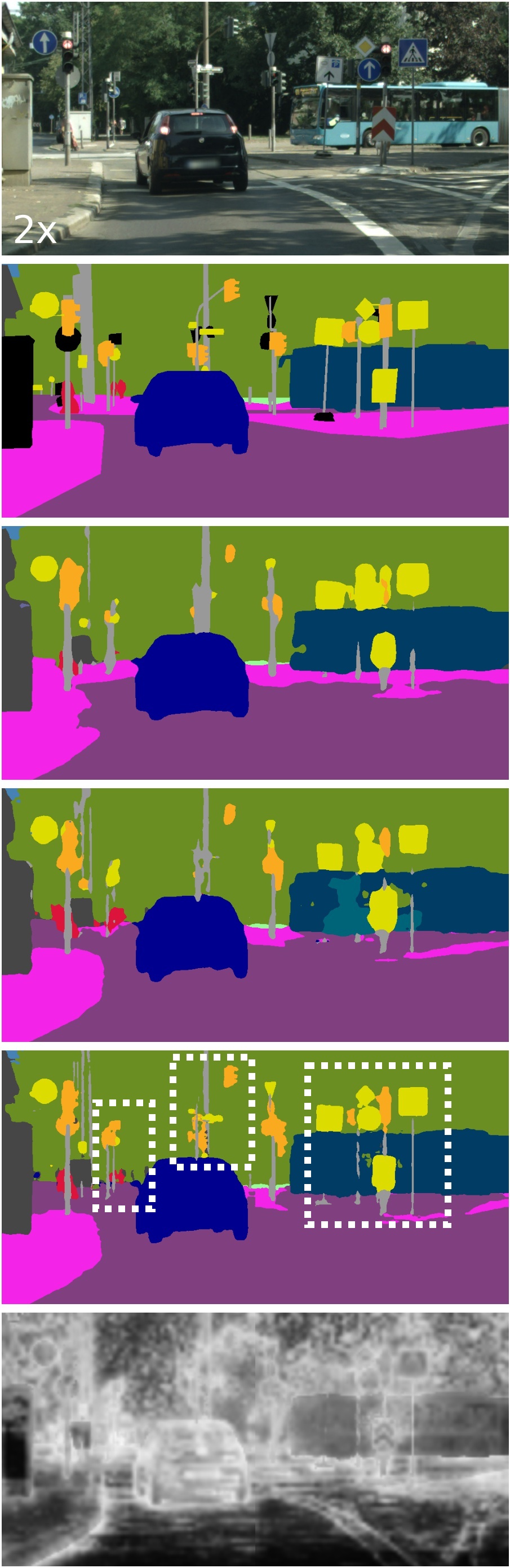}
\includegraphics[width=0.31\linewidth]{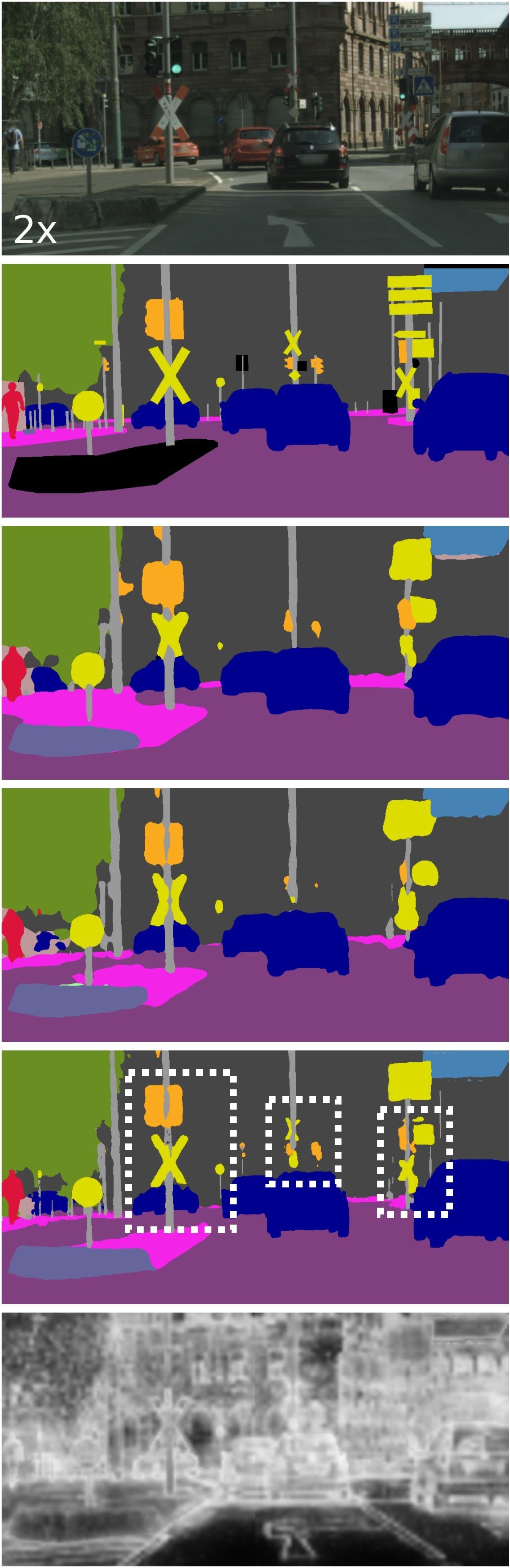}
\includegraphics[width=0.31\linewidth]{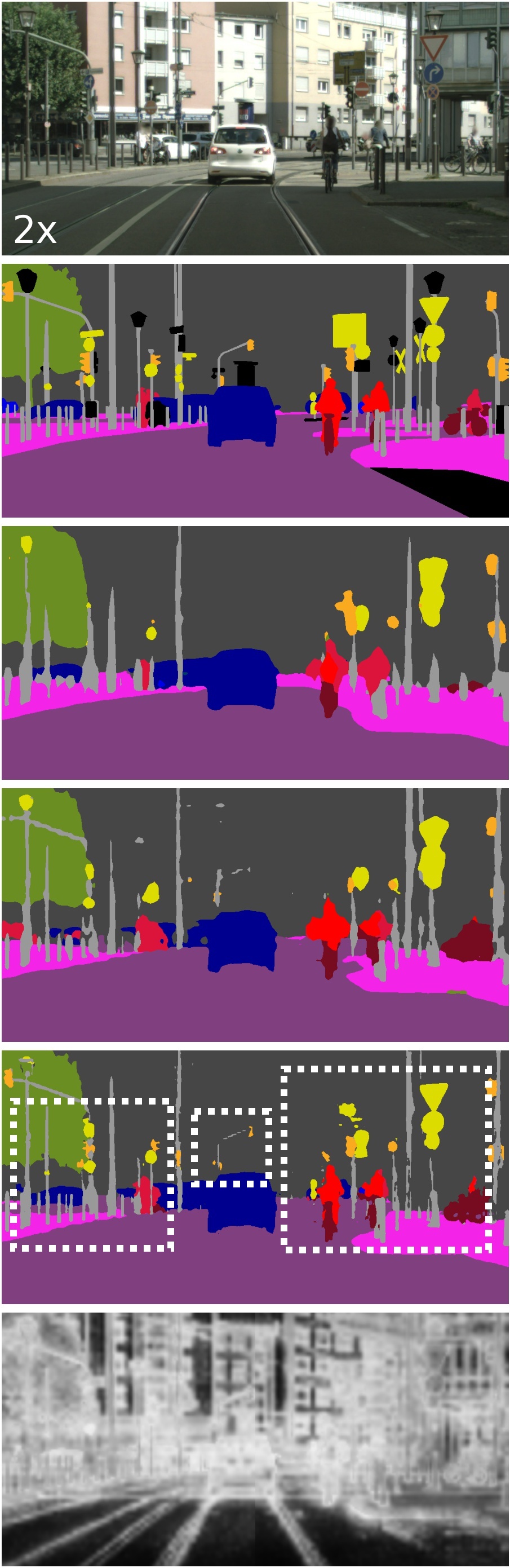}
\input{supp_preds/palette}
\caption{Example predictions showing a better recognition and finer segmentation details of small classes such as \emph{pole}, \emph{traffic light}, and \emph{traffic sign} on GTA$\rightarrow$Cityscapes. Some examples are zoomed in for better visibility of the details. The zoom factor is provided in the bottom left corner of each image.}
\label{fig:predictions_pole}
\end{figure*}

\begin{figure*}
\centering
\input{supp_preds/prediction_head}
\includegraphics[width=0.31\linewidth]{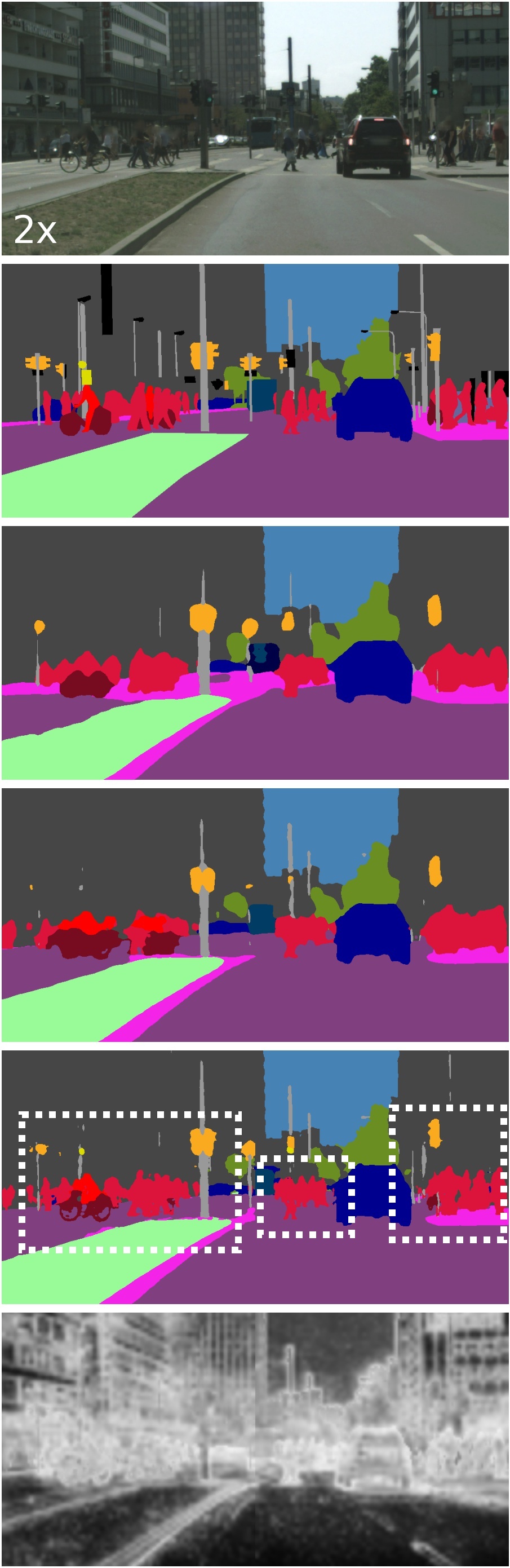}
\includegraphics[width=0.31\linewidth]{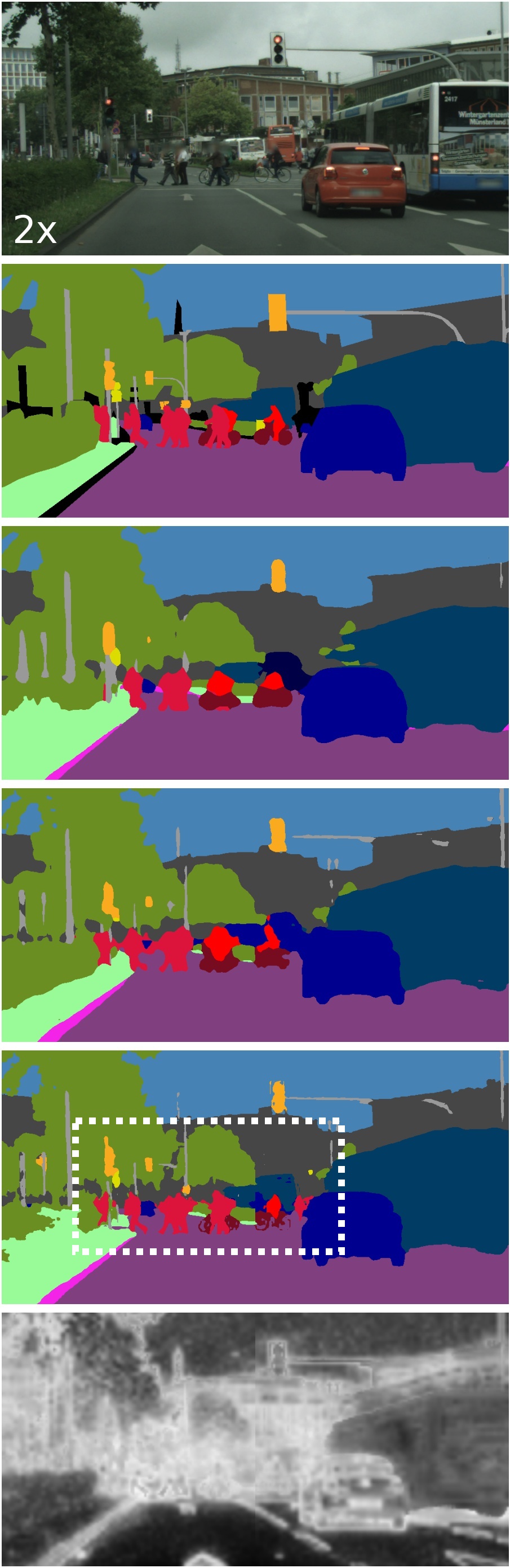}
\includegraphics[width=0.31\linewidth]{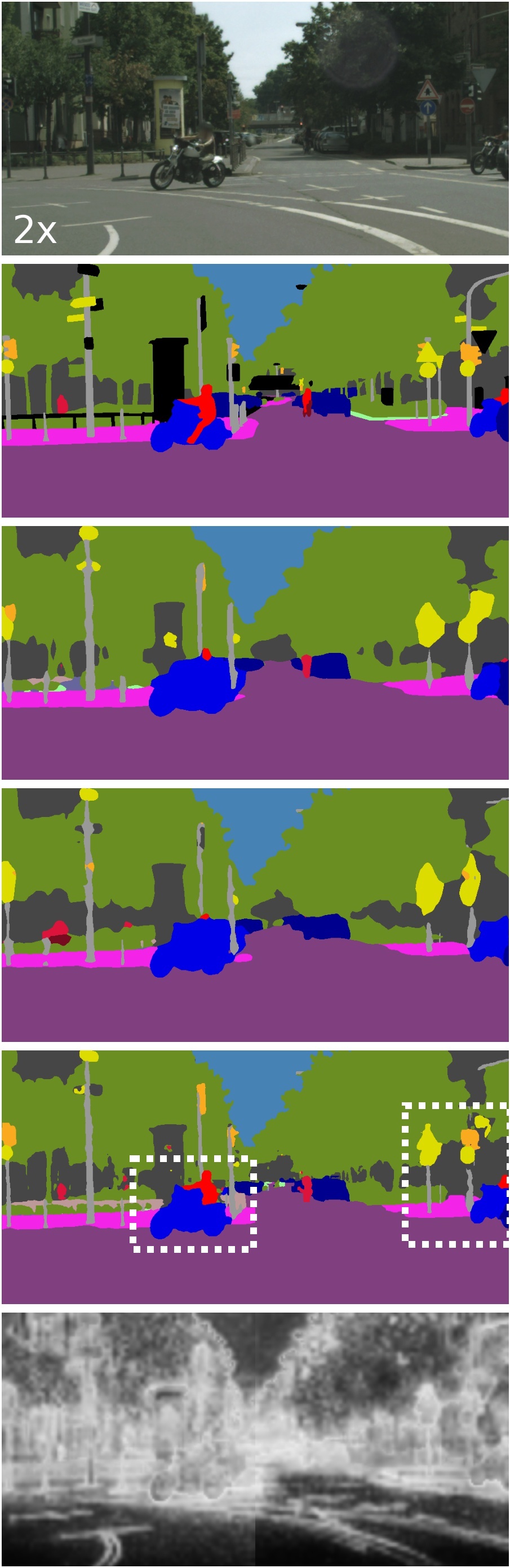}

\input{supp_preds/palette}
\caption{Example predictions showing a better recognition and finer segmentation of small distant classes such as \emph{pedestrian}, \emph{rider}, \emph{motorcycle} and \emph{bicycle} on GTA$\rightarrow$Cityscapes. Some examples are zoomed in for better visibility of the details. The zoom factor is provided in the bottom left corner of each image.}
\label{fig:predictions_rider}
\end{figure*}

\begin{figure*}
\centering
\input{supp_preds/prediction_head}
\includegraphics[width=0.31\linewidth]{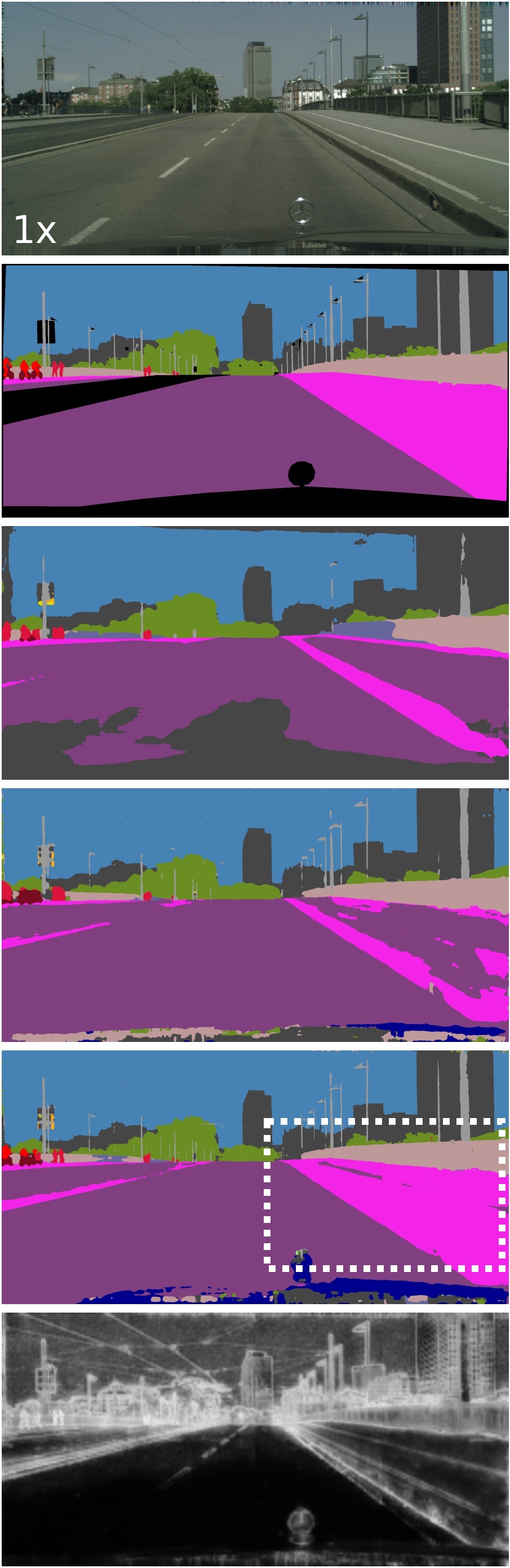}
\includegraphics[width=0.31\linewidth]{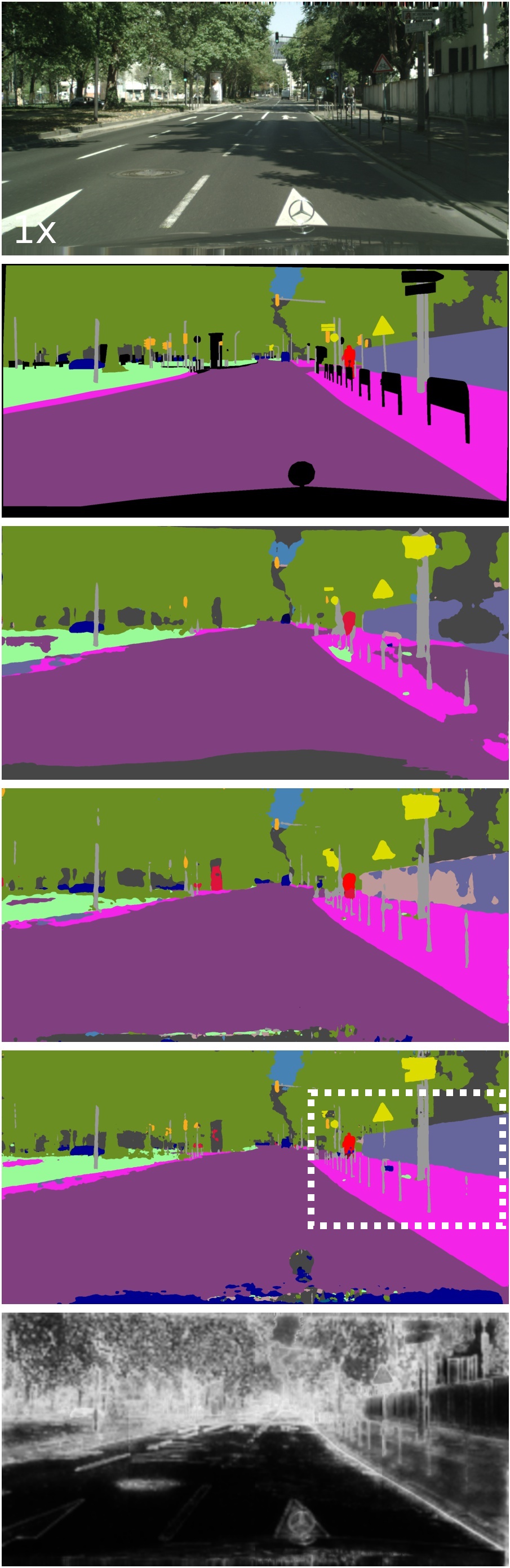}
\includegraphics[width=0.31\linewidth]{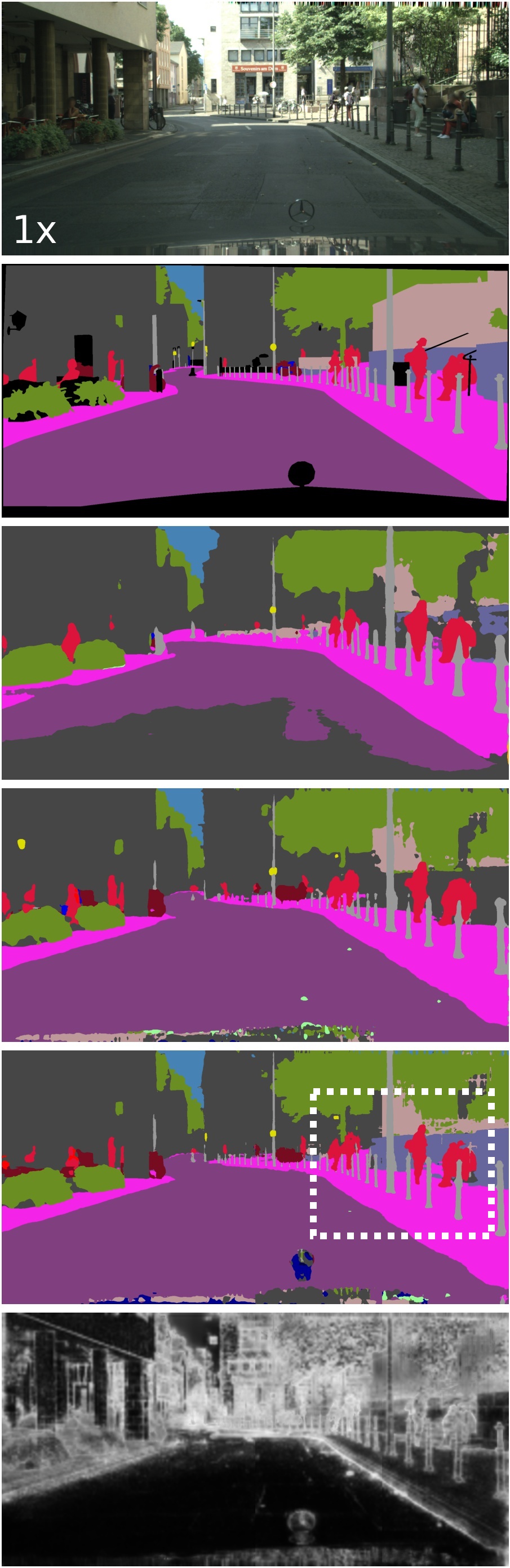}
\input{supp_preds/palette}
\caption{Example predictions showing a better recognition of difficult stuff classes such as \emph{sidewalk} and \emph{wall} on GTA$\rightarrow$Cityscapes. Some examples are zoomed in for better visibility of the details. The zoom factor is provided in the bottom left corner of each image.}
\label{fig:predictions_stuff}
\end{figure*}

\begin{figure*}
\centering
\input{supp_preds/prediction_head}
\includegraphics[width=0.31\linewidth]{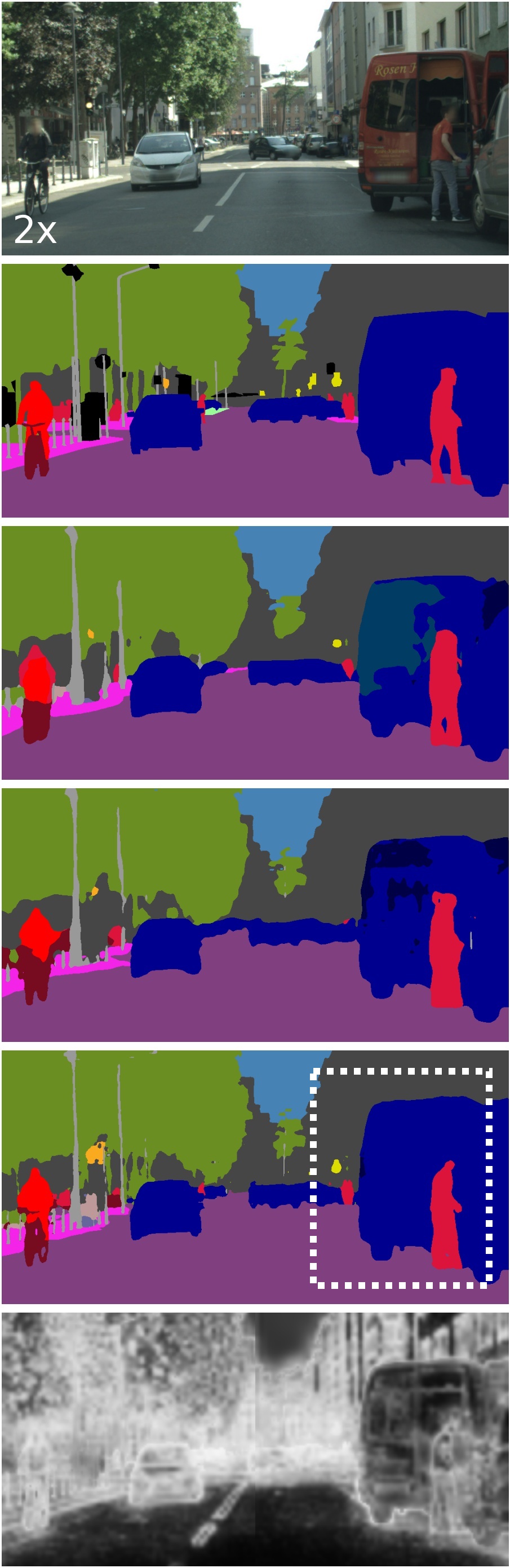}
\includegraphics[width=0.31\linewidth]{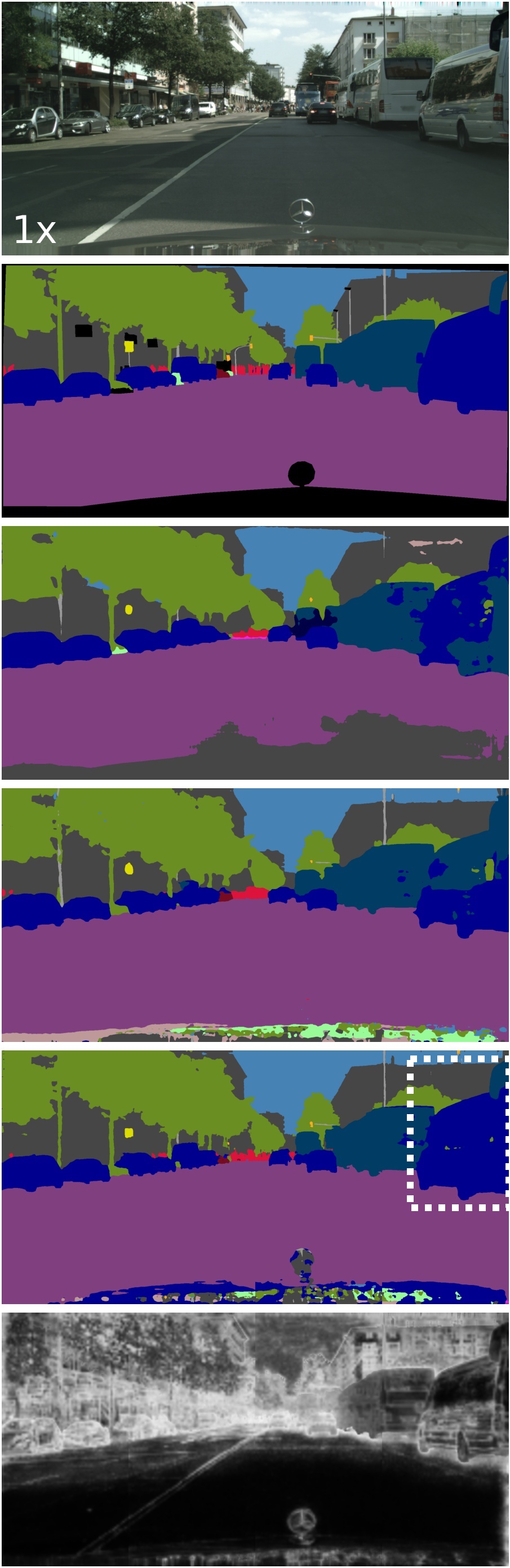}
\includegraphics[width=0.31\linewidth]{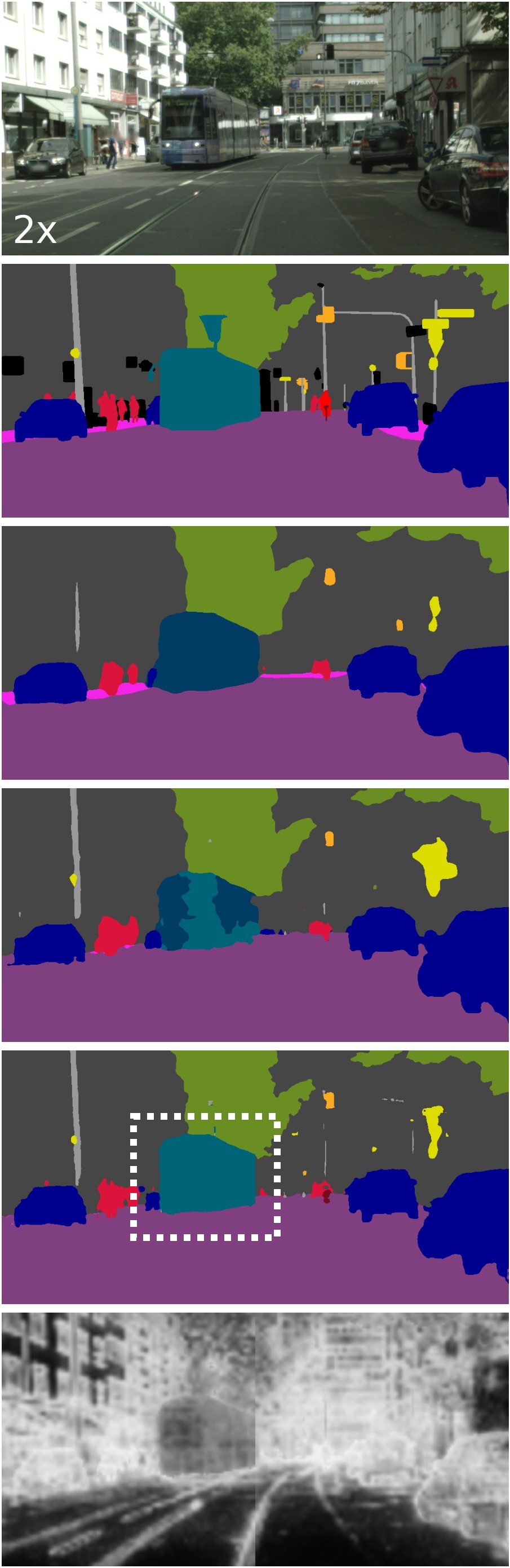}
\input{supp_preds/palette}
\caption{Example predictions showing a better differentiation of vehicle classes such as \emph{car}, \emph{truck}, \emph{bus}, and \emph{train} on GTA$\rightarrow$Cityscapes. Some examples are zoomed in for better visibility of the details. The zoom factor is provided in the bottom left corner of each image.}
\label{fig:predictions_vehicles}
\end{figure*}

\begin{figure*}
\centering
\input{supp_preds/prediction_head}
\includegraphics[width=0.31\linewidth]{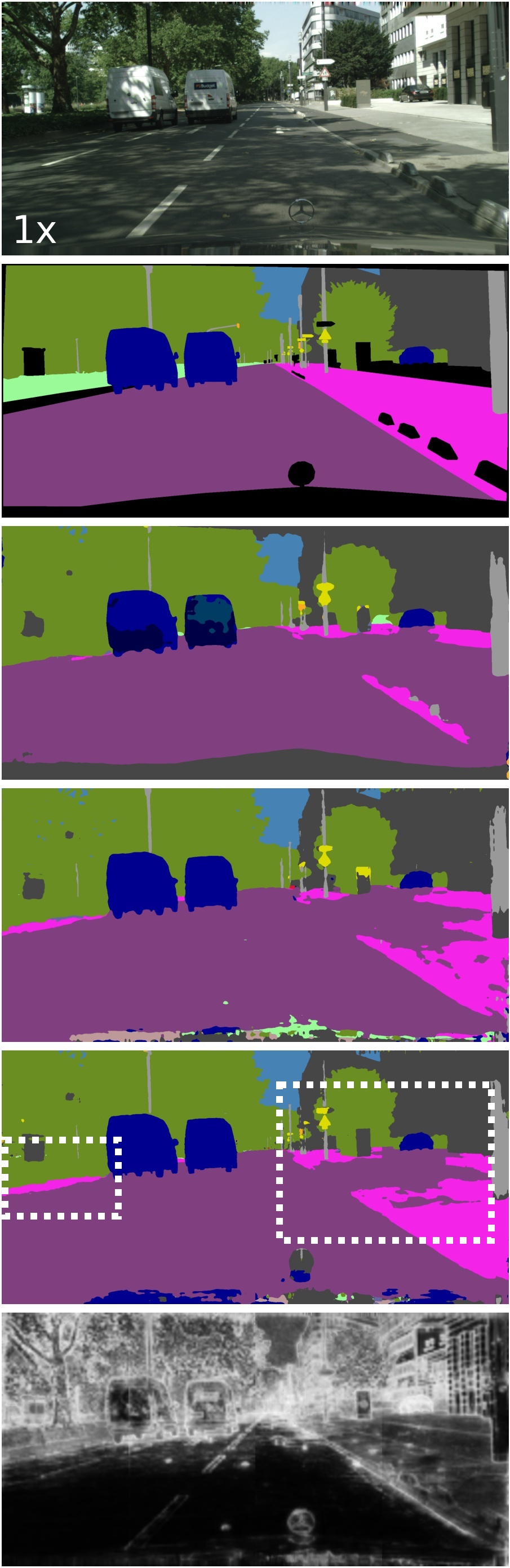}
\includegraphics[width=0.31\linewidth]{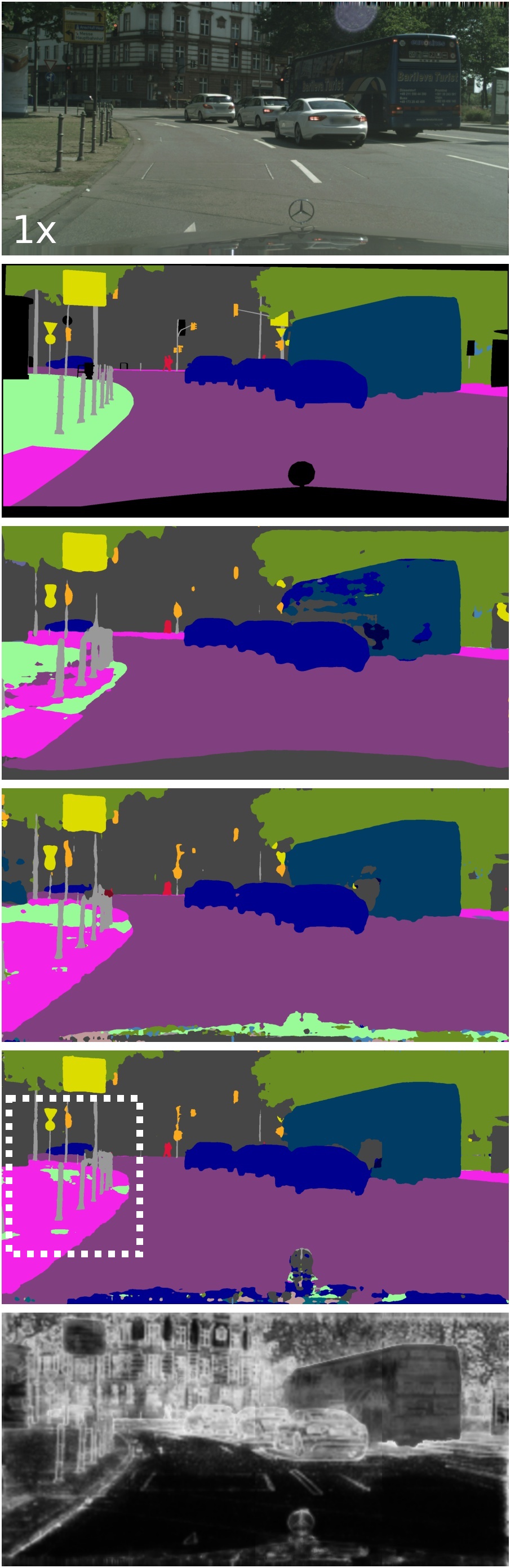}
\includegraphics[width=0.31\linewidth]{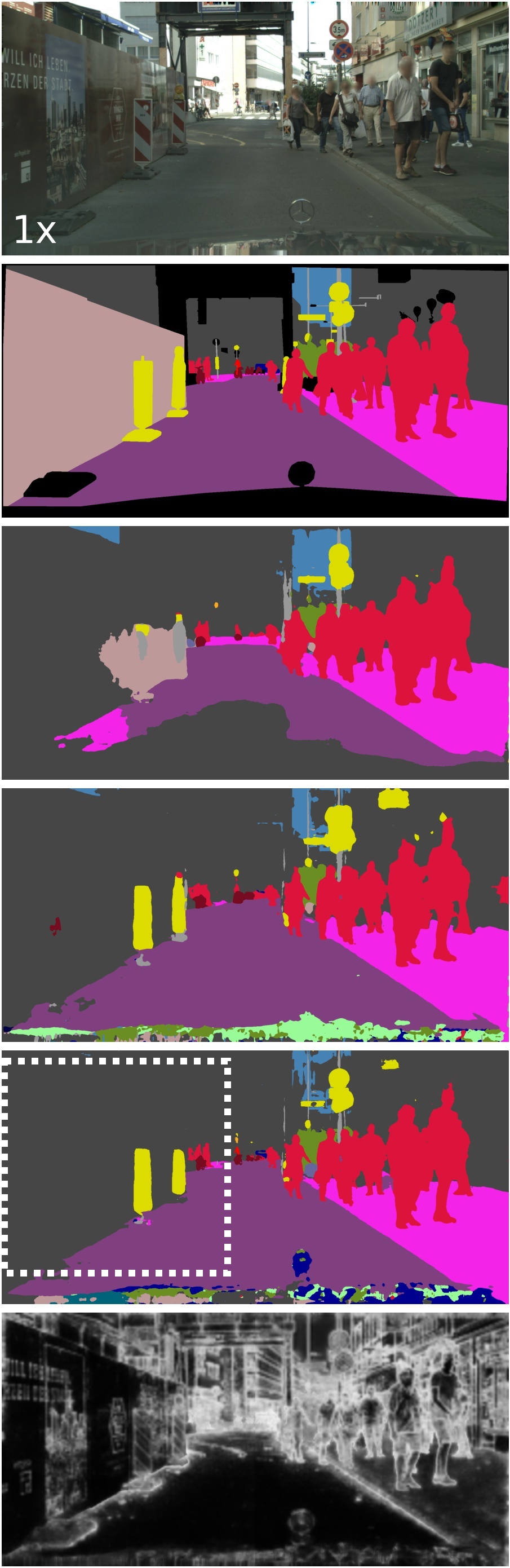}
\input{supp_preds/palette}
\caption{Failure cases of classes with a low UDA performance such as \emph{sidewalk}, \emph{terrain}, and \emph{fence} on GTA$\rightarrow$Cityscapes. Some examples are zoomed in for better visibility of the details. The zoom factor is provided in the bottom left corner of each image.}
\label{fig:predictions_failure}
\end{figure*}

In Fig.~\ref{fig:predictions_pole}-\ref{fig:predictions_failure}, we compare the predicted semantic segmentation of HRDA to the two strongest previous UDA methods from Tab.~\ref{tab:sota_gta}, namely ProDA~\cite{zhang2021prototypical} and DAFormer~\cite{hoyer2021daformer}. Further, we visualize the scale attention of HRDA as the weighted sum over the scale attention channels for each class with weight being the softmax of the segmentation prediction. White regions mean that HRDA focuses on the prediction from the HR input.

Overall, HRDA better recognizes small classes and segments finer details. This is especially the case for distant poles, traffic lights, and traffic signs (see Fig.~\ref{fig:predictions_pole}) as well as distant pedestrians, riders, bicycles, and motorcycles (see Fig.~\ref{fig:predictions_rider}). For these regions, HRDA uses the prediction from the HR input as can be seen in the HRDA scale attention (white encodes a focus on HR). Further, HRDA is able to better recognize difficult stuff classes such as sidewalk and wall (see Fig.~\ref{fig:predictions_stuff}) as well as to better distinguish different vehicle classes (see Fig.~\ref{fig:predictions_vehicles}). HRDA uses LR input for that purpose as can be seen in the HRDA scale attention (black encodes a focus on LR).

Even though HRDA sets new standards, UDA is still a challenging task. This can be observed for classes that are easy to confuse with others and that have a considerable domain gap such as sidewalk, terrain, or fence, which results in adaptation errors (see Fig.~\ref{fig:predictions_failure}).

\FloatBarrier

%% file: supp_preds/prediction_head.tex
\rotatebox{90}{%
{\scriptsize
\begin{tabularx}{11.7cm}{*{6}{Y}}
HRDA Att. & \textbf{HRDA} (Ours) & DAFormer~\cite{hoyer2021daformer} & ProDA~\cite{zhang2021prototypical} & G. Truth & Image \\
\end{tabularx}
}} %

%% file: supp_preds/palette.tex
\resizebox{\linewidth}{!}{%
\scriptsize
\setlength\tabcolsep{1pt}
{
\newcolumntype{P}[1]{>{\centering\arraybackslash}p{#1}}
\begin{tabular}{@{}*{10}{P{0.09\columnwidth}}@{}}
     {\cellcolor[rgb]{0.5,0.25,0.5}}\textcolor{white}{road} 
     &{\cellcolor[rgb]{0.957,0.137,0.91}}sidew. 
     &{\cellcolor[rgb]{0.275,0.275,0.275}}\textcolor{white}{build.} 
     &{\cellcolor[rgb]{0.4,0.4,0.612}}\textcolor{white}{wall} 
     &{\cellcolor[rgb]{0.745,0.6,0.6}}fence 
     &{\cellcolor[rgb]{0.6,0.6,0.6}}pole 
     &{\cellcolor[rgb]{0.98,0.667,0.118}}tr. light
     &{\cellcolor[rgb]{0.863,0.863,0}}tr. sign 
     &{\cellcolor[rgb]{0.42,0.557,0.137}}veget. 
     &{\cellcolor[rgb]{0.596,0.984,0.596}}terrain\\
     {\cellcolor[rgb]{0.275,0.510,0.706}}sky
     &{\cellcolor[rgb]{0.863,0.078,0.235}}\textcolor{white}{person} 
     &{\cellcolor[rgb]{1,0,0}}\textcolor{white}{rider} 
     &{\cellcolor[rgb]{0,0,0.557}}\textcolor{white}{car} 
     &{\cellcolor[rgb]{0,0,0.275}}\textcolor{white}{truck} 
     &{\cellcolor[rgb]{0,0.235,0.392}}\textcolor{white}{bus}
     &{\cellcolor[rgb]{0,0.392,0.471}}\textcolor{white}{train} 
     &{\cellcolor[rgb]{0,0,0.902}}\textcolor{white}{m.bike} 
     & {\cellcolor[rgb]{0.467,0.043,0.125}}\textcolor{white}{bike}
     &{\cellcolor[rgb]{0,0,0}}\textcolor{white}{n/a.}
\end{tabular}
}
}